\documentclass[conference]{IEEEtran}
\IEEEoverridecommandlockouts
\usepackage{cite}
\usepackage{amsmath,amssymb,amsfonts}
\usepackage{algorithmic}
\usepackage{graphicx}
\usepackage{textcomp}
\usepackage{hyperref}
\usepackage{xcolor}
\usepackage{ulem}
\usepackage[super]{nth}
\usepackage{subfigure}
\usepackage{bm}
\usepackage{multirow, booktabs}

\DeclareMathOperator*{\argmax}{argmax}

\makeatletter
\newcommand*{\rom}[1]{\expandafter\@slowromancap\romannumeral #1@}
\makeatother

\def\BibTeX{{\rm B\kern-.05em{\sc i\kern-.025em b}\kern-.08em
    T\kern-.1667em\lower.7ex\hbox{E}\kern-.125emX}}
\begin{document}

\title{Semantic-Based Active Perception for \\ Humanoid Visual Tasks with Foveal Sensors}

\author{\IEEEauthorblockN{João Luzio, Alexandre Bernardino, Plinio Moreno}
\IEEEauthorblockA{\textit{Instituto Superior Técnico} \\
Lisbon, Portugal \\
\href{mailto:joaoluzio14@tecnico.ulisboa.pt}{joaoluzio14@tecnico.ulisboa.pt}, \{\href{mailto:alexandre.bernardino@tecnico.ulisboa.pt}{alex}, \href{mailto:plinio@isr.tecnico.ulisboa.pt}{plinio}\}@isr.tecnico.ulisboa.pt}
}

\maketitle

\thispagestyle{plain}
\pagestyle{plain}

\begin{abstract} 
The aim of this work is to establish how accurately a recent semantic-based foveal active perception model is able to complete visual tasks that are regularly performed by humans, namely, scene exploration and visual search. This model exploits the ability of current object detectors to localize and classify a large number of object classes and to update a semantic description of a scene across multiple fixations. It has been used previously in scene exploration tasks. In this paper, we revisit the model and extend its application to visual search tasks. To illustrate the benefits of using semantic information in scene exploration and visual search tasks, we compare its performance against traditional saliency-based models. In the task of scene exploration, the semantic-based method demonstrates superior performance compared to the traditional saliency-based model in accurately representing the semantic information present in the visual scene. In visual search experiments, searching for instances of a target class in a visual field containing multiple distractors shows superior performance compared to the saliency-driven model and a random gaze selection algorithm. Our results demonstrate that semantic information, from the top-down, influences visual exploration and search tasks significantly, suggesting a potential area of research for integrating it with traditional bottom-up cues.
\end{abstract}


\section{Introduction}

\begin{figure}
\centering

\hspace{-10mm}\subfigure{\label{fig:regular}\includegraphics[width=0.72\columnwidth]{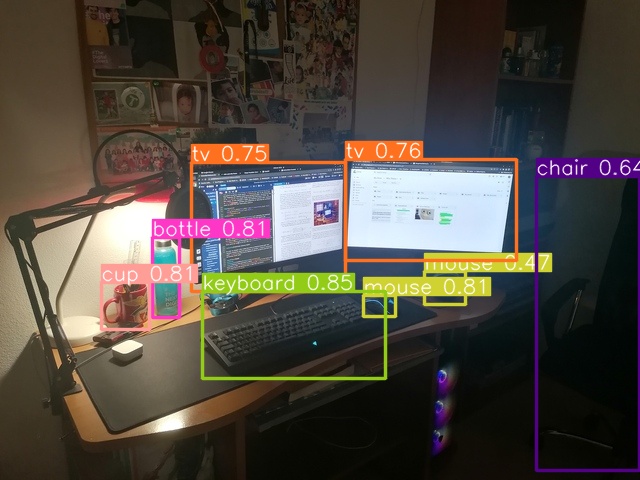}}

\hspace{10mm}\subfigure{\label{fig:foveated}\includegraphics[width=0.72\columnwidth]{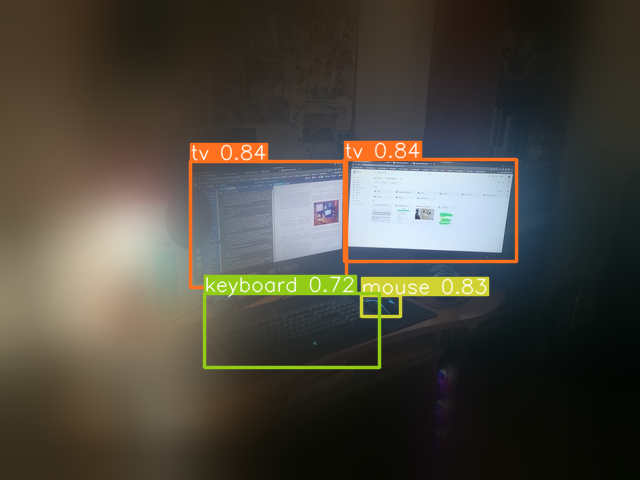}}  
\caption{Example of a regular image (top) that is artificially foveated (bottom) to emulate a visual field, together with the object predictions, outputted by an object detector, represented by their bounding-boxes and confidence scores.}
\label{fig:detection_example}
\end{figure}

Foveal vision \cite{foveal} and active perception \cite{activeperception} are the foundation of human visual cognition \cite{activevision}. Foveal vision reduces the amount of information to process during each gaze fixation, while active perception sequentially changes the gaze direction to the most promising regions of the visual field. Foveal vision is a means of improving the efficiency of brain computations, as, at each time, the brain has to process a smaller amount of information. In addition, foveal vision reduces the amount of information at the periphery of the visual field \cite{foveal} which reduces the saliency of the objects at those locations, e.g. \cite{activevision} (see Fig. \ref{fig:detection_example}). This challenges modern computer vision methods and models \cite{odreview} in terms of their ability to detect (localize and classify) such blurred objects.

Traditional cognitive models from psychology \cite{trad_sal} state that the guiding mechanism for human visual attention depends on a range of distinct types of conspicuity features, which are combined in a saliency map that highlights regions of interest. Modern approaches consider visual models that can extract bottom-up \cite{saliency}, top-down \cite{topdown}, or both types of features \cite{ivsn} to actively perform typical human visual tasks. 

Recent work, developed by Dias et al. \cite{main}, explores the possibility of combining the rich semantic information provided by modern object detectors with active perception, to perform scene exploration in foveal images. When exploring a scene, the goal is to accurately map the semantic content available in a visual field through successive premeditated gaze shifts.
The main challenges associated with this humanoid task are the ability to cope with the reduced resolution of the visual information available in the peripheral region and the accumulation of semantic information in successive gaze shifts \cite{predvisualfix} to improve the agent's belief in the presence of objects from a set of known classes. Both challenges are addressed using Bayesian methods \cite{uncertainty} and proper characterization of the uncertainty that is correlated with foveal image degradation.

In \cite{main} the semantic data generated by the object detectors are interpreted as Dirichlet random variables \cite{compound}, which enable proper data fusion \cite{kaplan} and calibration \cite{calibration}. Active perception methods are then formulated as determining gaze locations that minimize some measure of uncertainty. 
In this work, we consider applying the same methodology \cite{main} to another visual task performed by humans, namely visual search \cite{visualsearch}, where the objective is to set the gaze directly upon a region that contains an instance of a pre-defined target class. For this reason, the relevance of gaze selection is amplified given the greedy target-oriented nature of this cognition-related task.

In summary, the objective of this work is to establish how well the novel semantic-based active perception model \cite{main} is able to complete visual tasks that humans routinely perform. For this purpose, we compare its accuracy with a biologically inspired attention model \cite{saliency}, based on studies of human visual cognition \cite{trad_sal}, which attempts to mimic both the behavior and the neuronal architecture of the primate visual system. Following an appropriate experimental setup, we confirm that the semantic model is applicable in the context of visual search. More specifically, we can conclude that the semantic model significantly outperforms both a random gaze selection algorithm and the selected bottom-up saliency-based attention model \cite{saliency} when considering the experimental results obtained while performing both visual search and exploration tasks.

The outline of this document is as follows. In Section \ref{sec:back} we present a compilation of key concepts that are fundamental to our work. Then, in Section \ref{sec:methods} we describe the complete methodological apparatus that involves the semantic-based foveal active perception, to be applied during the experimental phase.  In Section \ref{sec:exp}, we present an overview of the experimental setup along with a comparison of the most relevant results obtained during the experiments, upon completion of multiple visual search tasks. A comparison of different approaches to active perception is established. Finally, in Section \ref{sec:conclusion} we highlight the most important contributions of this work and present the final remarks.

\section{Background and Related Work}\label{sec:back}

\subsection{Humanoid Visual System}

Despite being often compared to a photographic camera, the human eye's processing capacity across the visual field is not homogeneous. Several anatomical properties of the eye are, for example, correlated with the presence of gaps \cite{foveal} in sensory information. Due to these anatomical properties, the processing of visual signals varies quite dramatically across the visual field. 
Therefore, it is important to distinguish between the center of the visual field, known as the fovea, and an outer region, known as the periphery. Foveal vision ensures maximum acuity and contrast sensitivity in a small area around the gaze position \cite{foveal}, while peripheral vision allows for a large field of view, although it presents lower resolution and contrast sensitivity, as well as higher object positional \cite{activevision}.

A classical approach to generate artificially foveated images from regular Cartesian images is to use methods based on log-polar transformations \cite{logpolar} to generate cortical maps. A cortical map is a topographic representation of sensory or motor information in the brain, specifically found in the cerebral cortex. A log-polar space is the most appropriate approach to generate cortical images \cite{siebert}, modeling with reasonable fidelity the mapping observed in the primate visual cortex.
Other approaches apply Cartesian foveal geometry \cite{biofov} to avoid the nonlinearities \cite{logpolar} associated with log-polar spaces. 
In our methodology, we use an artificial foveal system proposed by Almeida et al. \cite{fovsys} to foveate regular Cartesian images. 
\begin{figure}
\centering
\subfigure{\label{fig:nonfov}\includegraphics[width=0.32\columnwidth]{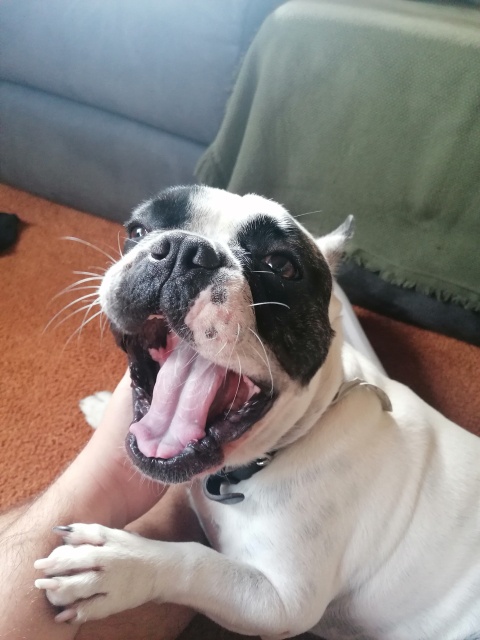}} 
\subfigure{\label{fig:fov100}\includegraphics[width=0.32\columnwidth]{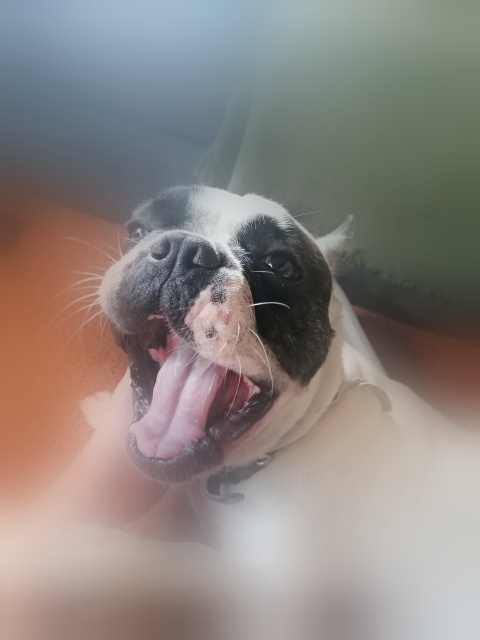}}  
\subfigure{\label{fig:fov50}\includegraphics[width=0.32\columnwidth]{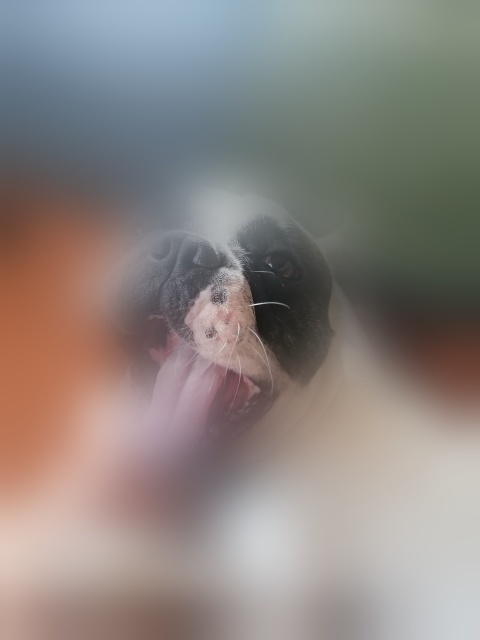}}  
\caption{Example of a Cartesian image (left) to which the artificial foveal system, proposed by Almeida et al. \cite{fovsys}, has been applied. There are presented two foveal images, with wider (center) and narrower (right) foveal dimensions.}
\label{fig:foveal_reconstruction}
\end{figure}
This system operates on the basis of a combination of Gaussian and Laplacian pyramids, taking advantage of space-variant low-pass spectral filters to introduce different levels of blur on the distinct regions of the image, precisely as depicted in Fig. \ref{fig:foveal_reconstruction}. One advantage of this foveation model is that it is fast \cite{milicio}, making it applicable in a real-time context.

\subsection{Semantic Object Detection}

In general, object detection models aim at classifying existing objects in any provided image, typically identified by a rectangular box (bounding-box) and an associated vector of scores, typically normalized to be interpreted as class posterior probabilities. As described in \cite{odreview}, there are three main stages in traditional object detection models: selection of informative regions, extraction of features, and classification of objects. Several approaches follow this pipeline when solving object detection, each of them unique in some particular aspect. Object detection frameworks can mainly be divided into two types: The first approach (two-stage frameworks) follows the traditional pipeline of stages and consists of first generating region proposals and then classifying each proposal into the relevant object categories (e.g. R-CNN \cite{rcnn}). In the second approach (one-stage frameworks), object detection is viewed as a regression or classification problem, where both categories and locations are determined directly through the use of a unified framework (e.g. SSD \cite{ssd}, YOLO \cite{yolo_og, yolo}). These frameworks have been propelled by the overwhelming power of Convolutional Neural Networks (CNNs) given their ability to process features that are hierarchy extracted from images. Moreover, while traditional object detection methods (e.g. \cite{rcnn}) involve separate stages for region proposal and object classification \cite{odreview}, more modern transformer-based approaches (e.g. EVA \cite{eva}, DETR \cite{detr}) aim to perform both tasks in a single step, leveraging the attention mechanism in transformers for capturing contextual information across the image.

\subsection{Human Cognitive Mechanisms}

Biological organisms use selective visual attention mechanisms to process single portions of the available visual information while disregarding the remainder. This enables these organisms to efficiently perceive and handle the limited neural resources in the brain, which are allocated to vision.
Furthermore, visual attention covers all factors that influence information selection mechanisms \cite{activevision}, whether guided by visual stimuli, which we refer to as bottom-up features, or by task-related expectations, denominated as top-down features. The Attentional Engagement Theory (AET), also known as Similarity Theory, asserts that stimulus objects are represented at a level of perceptual description, where top-down object representations \cite{visualsearch} compete with each other to enter visual short-term memory. These representations enhance the target-distractor dissimilarities when the competition is biased toward certain target objects, such as in a visual search task. 

\begin{figure}
\centering
\includegraphics[width=\columnwidth]{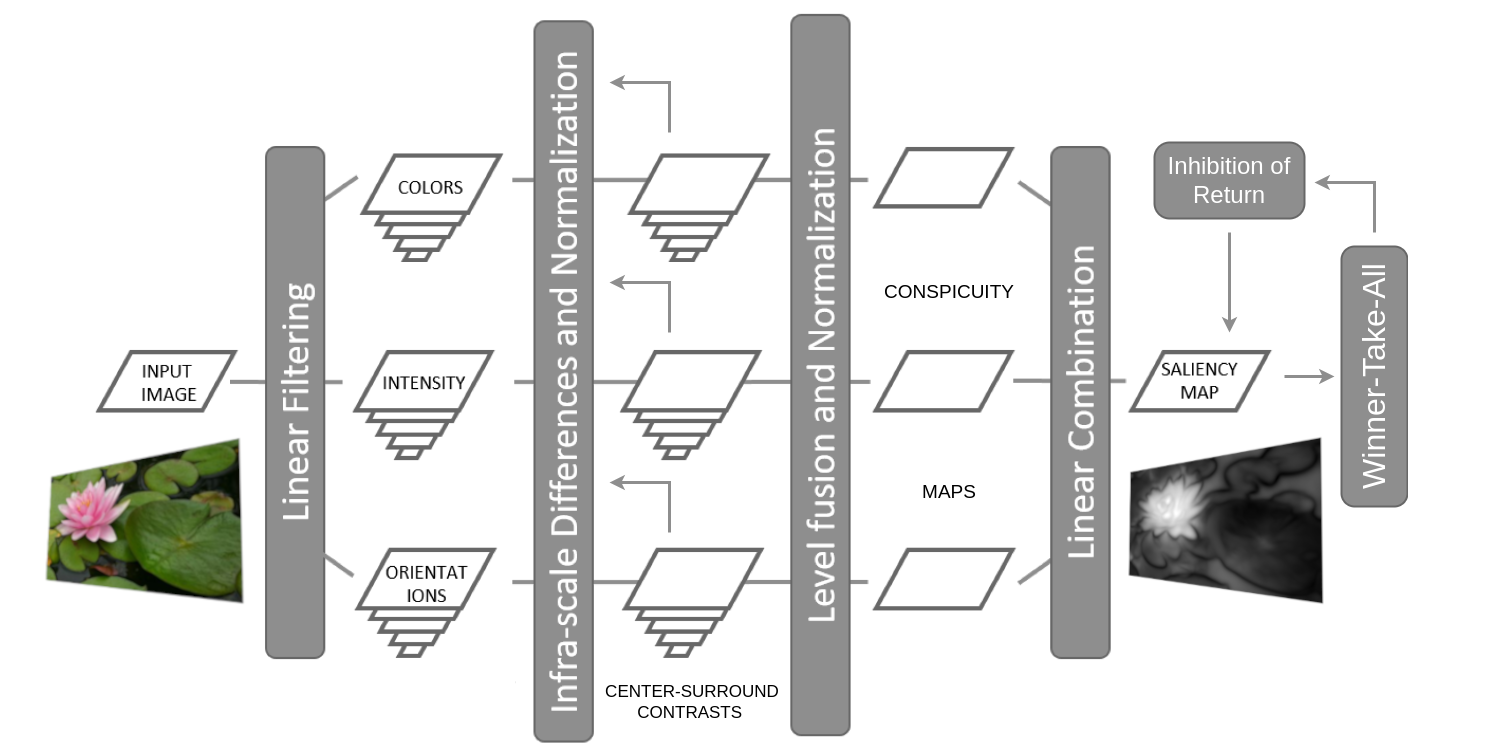}
\caption{Simplified version of the Itti-Koch attention system \cite{trad_sal}, inspired by both the neuronal architecture and behavior of the early primate visual system.}
\label{fig:attention}
\end{figure}

\begin{figure*}
\centering
  \includegraphics[width=0.9\textwidth]{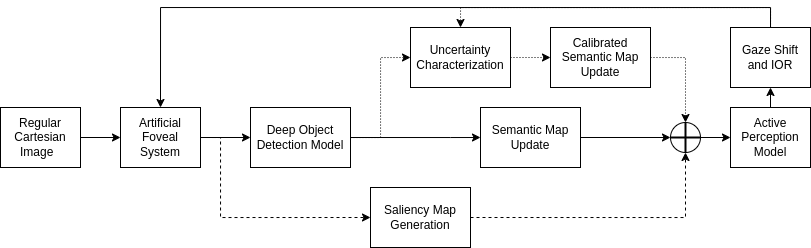}
  \caption{Our general methodological approach (based on \cite{main}) to semantic visual search and scene exploration tasks. An image is foveated on some initial image location, to simulate the human visual sensor. The foveal image is fed to an object detection model that may generate multiple bounding-boxes and the respective categorical scores. This information is used to update a world-fixed semantic map. This can be done in two ways: using the raw scores of the object detector (solid line) or with scores calibrated according to the effects of the foveal image characteristics (dotted line), since increased blur in the periphery will increase uncertainty to the classification scores. Alternatively, we also tested the use of a traditional saliency map approach to predict the next best focal point. The image is then foveated at the new selected focal point, simulating a saccade. A clasical Inhibition of Return method (IOR) \cite{trad_sal} is applied to prevent returning to a previously visited location. The process is repeated until some terminal condition is met.}
  \label{fig:general}
\end{figure*}

A classical neural architecture proposed by Itti et al. \cite{trad_sal} proposes a pure bottom-up approach to generate a saliency master map that is constructed in the posterior parietal cortex of primates (illustrated in Fig. \ref{fig:attention}). The feature integration theory (FIT) \cite{visualsearch}, which is a well-known cognitive theory, states that different features are processed in distinct brain regions. In line with the guidelines set out in the FIT \cite{trad_sal}, Itti's model utilizes color and intensity variations to produce several layered feature maps that depict the differences seen in prominent areas. A winner-take-all (WTA) decision model selects the most salient region out of a master map that linearly combines conspicuity maps in which different feature layers are fused. In addition, an inhibition of return (IOR) mechanism ensures that when transversing the visual field to complete a visual task, previously visited regions are avoided by the WTA mechanism.

The traditional Itti-Koch attention model \cite{trad_sal} inspired multiple modern approaches to human visual cognition, and among those is VOCUS2 \cite{saliency}, an improved and modernized version of this classical model. This model comprises multiple necessary adaptations that maximize the performance of the model, with special emphasis on the scale-space, enabling a ﬂexible center-surround ratio definition. VOCUS2 \cite{saliency} is not only coherent in structure, but also fast and, as a typical bottom-up conspicuity model, performs saliency computations solely at the pixel level.
The Invariant Visual Search Network (IVSN) \cite{ivsn} is an alternative model that takes advantage of the immense power of CNNs to extract top-down high-level features from a visual scene, in order to perform a visual target search. Similarly to the traditional attention model \cite{trad_sal}, the proponents of IVSN consider both WTA and IOR mechanisms when iteratively selecting the most promising regions from the field of view.   

\subsection{Active Perception and Attention Models}

Active perception is a type of active learning \cite{activeperception} in which an agent gathers information from sensors and combines the current state of information with prior knowledge of the world to determine the next action to be taken. This approach can be used with various types of sensors and stimuli (for example, tactile, auditory, and olfactory) \cite{activeperception}, yet our study focuses on the use of visual sensory information to perform search and exploration tasks. In the context of these visual tasks, the next move consists of a gaze shift toward a different focal point. Therefore, in this context, the active decision consists of selecting the most promising region to where the gaze should then be shifted.
To guide the selection process, Figueiredo et al. \cite{apfov} considered the usage of acquisition functions, such as the probability of improvement or the expected improvement, to choose the point that maximizes a task-specific function, through stochastic optimization. This is accomplished through the application of appropriately pre-defined metrics, such as the classification entropy or the categorical probability of a particular class.

Visual attention covers all factors that influence information selection mechanisms \cite{activevision}, whether guided by intrinsic aspects of the visual stimuli (bottom-up features) \cite{trad_sal,context} or by task-related expectations (top-down features) \cite{main,ivsn}. On the one hand, the traditional saliency-based attention mechanism, proposed by Itti et al. \cite{trad_sal}, is the baseline model that inspired multiple modern approaches to human visual cognition, such as VOCUS2 \cite{saliency}, an improved and modernized version of this classical model, and low-level models that explore the relevance of contextual guidance \cite{context} for eye movements and human attention. On the other hand, top-down models, such as the Invariant Visual Search Network (IVSN) \cite{ivsn} and SaltiNet \cite{saltinet}, take advantage of the immense power of Convolutional Neural Networks (CNNs) to extract top-down high-level features, to perform visual search \cite{visualsearch} in the presence of multiple distractors. Other advanced models take advantage of Deep Markov Models (ScanDMM \cite{scandmm}), spatial priority and scanpath networks (DeepGazeIII \cite{kummerer}), generative adversarial networks (PathGAN \cite{pathgan}), object detection models within a probabilistic framework \cite{main}, and visual transformers (HAT \cite{transformers}) to model task-specific human visual scanpaths, using different types of visual inputs (e.g. Cartesian, foveal, 360-degree images) and suitable 2D/3D spatial representations \cite{apfov,objsearch}. A significant number of new fixation prediction models are published every year. These models can be evaluated with properly maintained benchmarks, such as the MIT/Tuebingen Saliency Benchmark \cite{benchmark}, which hosts specialized saliency datasets (COCO-Search18 \cite{cocosearch18}, MIT300 \cite{mit300}, CAT2000 \cite{cat2000}) for visual tasks such as free-viewing and visual search, using well-established metrics \cite{benchmark}. Despite the considerable advancements that resulted from training modern deep learning models for visual attention \cite{sota}, there is still a large margin for improvement when it comes to the integration of these models with biologically inspired mechanisms, derived from neurological and psychological principles and theories, such as hard attention \cite{siebert,biofov,saccader}, low-level bottom-up features \cite{trad_sal,saliency} and contextual information \cite{context}. With this integration of concepts, through the combination of computer vision, visual cognition, and possibly Large Language models (LLMs) or Visual Language models (VLMs) that extract human knowledge on high-level relationships \cite{llm}, emerges the opportunity to develop gaze fixation models that better mimic human visual-cognitive behaviors.

Recently developed models that consider either uniform Cartesian \cite{topdown,ivsn} or foveal \cite{milicio,siebert,biofov} visual systems, make use of CNNs to extract and process bottom-up, top-down, or both types of attentional features, which can be used to spot regions of interest over the entire visual field. Alternative approaches \cite{objsearch,main} exploit the rich semantic information outputted by modern deep object detection models, building context grids that map the information collected from multiple object predictions distributed throughout the field of view. Some other approaches (e.g. \cite{objsearch,grotz}) even consider modeling the full extent of humanoid behavior, allowing not only for eye movements but also full head and neck mobility.

\section{Methodology}\label{sec:methods}

The fundamental concept supporting our methodology, illustrated in Fig. \ref{fig:general}, is based on the idea that for most human activities, especially those involving visual tasks, it is highly beneficial to utilize previous sensory data \cite{activeperception} to plan for the next action. Such a strategy can efficiently minimize the number of actions (i.e. gaze shifts) necessary to complete the task at hand. Following the approach proposed by Dias et al. \cite{main} we leverage the semantic information obtained by the state-of-the-art deep object detection models. 
This data is essential for pinpointing specific areas of importance within a scene, enabling the agent to make more informed decisions on where to focus its gaze next \cite{activeperception}, particularly when engaging in human-like visual activities like visual search \cite{visualsearch}, and utilizing artificial foveal vision \cite{fovsys} to replicate the human field of vision.
\subsection{Outline of the Model}
\subsubsection{Spatial Representation}
\begin{figure}
\centering
\includegraphics[width=\columnwidth]{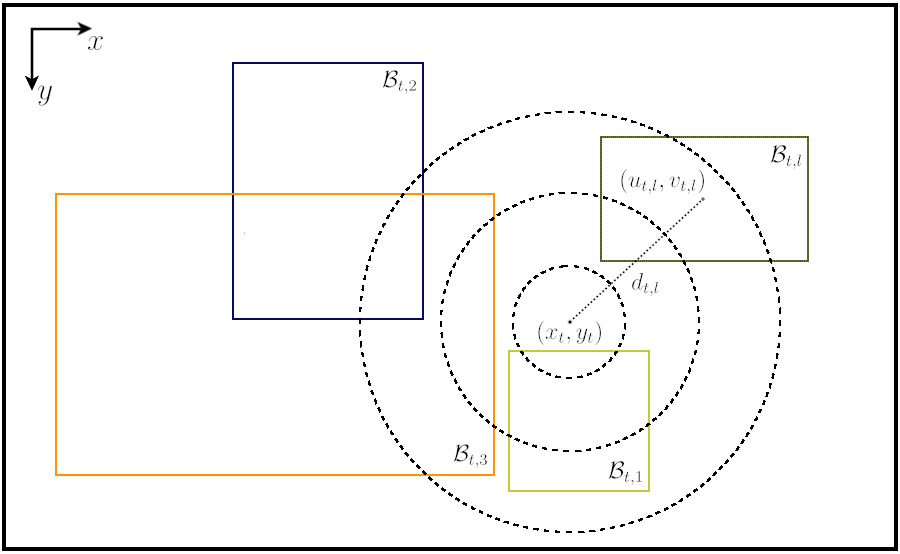}
\caption{Illustration of a foveal scene, fixed on a focal point, from which an arbitrary detector outputs $L_t$ object predictions. The distance $d_{t,l}$ between the center of a bounding-box $\mathcal{B}_{t,l}$ and the center of the fovea is also represented.}
\label{fig:map}
\end{figure}

In the context of robotic vision, there are multiple ways \cite{apfov} to represent the geometry of the surrounding environment (e.g. voxel grids, point clouds). In our approach, we consider a fixed field of view, meaning that the agent can only perform simple ocular movements (i.e. saccades) without dynamically changing the configuration and the boundaries of the visual field. Taking this assumption into account, we select a typical two-dimensional Cartesian representation (namely, an occupancy grid \cite{apfov}) that spatially encodes the surrounding environment into $\mathbf{x} = (x,y)$ coordinates. This representation is illustrated in Fig. \ref{fig:map} and is used to implement the \textit{Semantic} and \textit{Saliency} maps of Fig. \ref{fig:general}. Moreover, in our approach, we consider and assume that the visual field has a static configuration \cite{main}, where the dispositions of the objects contained in it are immutable.

\begin{figure*}
\centering
  \includegraphics[width=\textwidth]{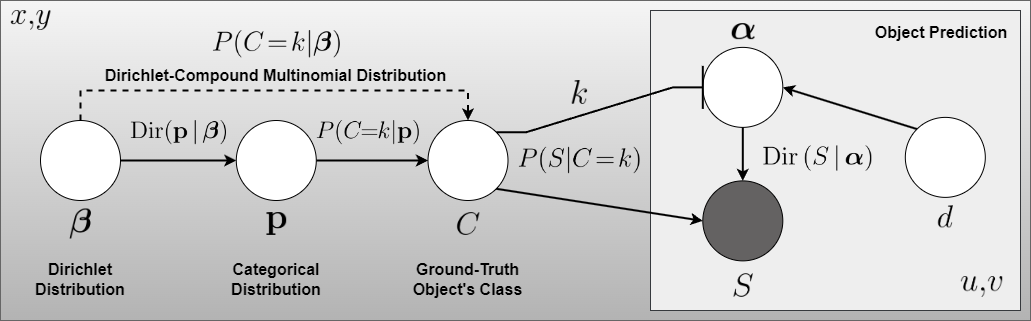}
  \caption{Representation of the dependencies between the variables that participate in the methodology, as proposed by Dias et al. \cite{main}, in the form of a Bayesian network, using plate notation, explaining both semantic content mapping and foveal score calibration techniques. Essentially, we attach a Dirichlet probability distribution, with parameters $\boldsymbol{\beta}$, to each pair of coordinates $(x,y)$ of the context grid, which, in the context of Bayesian inference, serves as the prior for a categorical distribution $\mathbf{p}$ that represents the confidences in the presence of each class $k\in\mathcal{C}$. Furthermore, we simplify the notation by defining the actual posterior distribution for the ground truth class $C$ through a Dirichlet compound multinomial distribution \cite{compound}, as detailed in Section \ref{sec:fusion}.}
  \label{fig:bayesnet}
\end{figure*}
\subsubsection{The Semantic Model}


We assume that the agent can observe, at each spatial location $(x,y)$, one object of possible $K$ classes. The absence of objects is represented by the class $k=0$ (background class). At a given time stamp $t$, with the gaze fixed on a certain region with $\mathbf{f}_t = (x_t,y_t)$ coordinates, the object detection model \cite{odreview} outputs a set of object predictions $\mathcal{I}_t$, composed of $L_t$ detections. As depicted in Fig. \ref{fig:map}, each detection $I_{t,l} \!\in \mathcal{I}_t$, consists of a bounding-box $\mathcal{B}_{t,l}$ and a normalized score vector\footnote{In case the detector's output \eqref{eq:scores} is not normalized, the scores can be normalized by dividing each $s_{t,l,k}$ by $\sum_{i=0}^K s_{t,l,i}$.} $S_{t,l}$ of dimension $K+1$ (including the background class), that is,
\begin{equation}
    S_{t,l} = \left( s_{t,l,0}, s_{t,l,1}, \thinspace \dots \thinspace,  s_{t,l,K} \right), 
    \label{eq:scores}
\end{equation}
\noindent where $0\leq s_{t,l,k} \leq 1, \forall k \in \mathcal{C}$ and $\sum_{i=0}^K s_{t,l,i} = 1$. $\mathcal{C} \subseteq \left[0,\dots,K \right]$ is the set of possible classes and $l=1,\dots,L_t$. 
Through probabilistic inference, we want to create a map that aggregates all the observations obtained in the past ($\mathcal{I}_{1:t}$).
The fused information should represent the semantic knowledge accumulated about the object classes considered $C^{\mathbf{x}} \in \mathcal{C}$, existing in the possible fixation points $\mathbf{x}$ in the context grid  \cite{predvisualfix}
\begin{equation}
    \mathbf{M}_t (\mathbf{x}) \!=\! \left[ P \left( C^{\mathbf{x}} \!=\! 0 \mid \mathcal{I}_{1:t} \right), \dots , P \left( C^{\mathbf{x}} \!=\! K \mid \mathcal{I}_{1:t} \right) \right]^T \!\!,
    \label{eq:semanticmap}
\end{equation}
\noindent where $P \left( C^{\mathbf{x}} = k \mid \mathcal{I}_{1:t} \right)$ represents the posterior probability of class $k \in \mathcal{C}$, given all past locally accumulated semantic information. 

\subsubsection{Semantic Map Update}
The most straightforward way to update the semantic map \eqref{eq:semanticmap} uses the raw classifier score vectors \eqref{eq:scores}, as conveyed by the straight line path in Fig. \ref{fig:general}. We use classifier fusion methods inspired in \cite{kaplan}, which will be presented in Section \ref{sec:fusion}. 

A more sophisticated way to update the semantic map, illustrated by the dotted line path in Fig. \ref{fig:general}, takes advantage of the fact that observations at different eccentricities in the fovea yield different levels of uncertainty.  Object detection models that have been pre-trained on a large set of regular Cartesian images are completely unaware of the existence of radial blur levels \cite{foveal} that characterize foveal images \cite{activevision}. This distortion imposed by a foveal visual system is responsible for a significant portion of the data-induced uncertainty observed in the generated detections. A fine-tuned calibrated probabilistic classifier of $(K\!+\!1)$ classes, similar to those incorporated within contemporary deep object detection models such as \cite{calibration}, can correctly quantify the level of uncertainty inherent in its instance-wise predictions. For this reason, we apply an efficient technique for uncertainty calibration \cite{uncertainty}, considering scores \eqref{eq:scores} as random variables sampled from a Dirichlet distribution $S \sim \mathrm{Dir} (\boldsymbol{\alpha})$ and use the Bayes theorem to find the class posterior distribution $P(C=k \vert S)$, as presented in Section \ref{fom}. In Section \ref{sec:fom} we present a calibration method that accounts for the space-variant distortion that accompanies the prediction of each object. 

\subsubsection{Active Perception}
Finally, in accordance with the block diagram presented in Figure \ref{fig:general}, an active perception model extracts the information contained in one of the available maps to predict the next state of \eqref{eq:semanticmap} and select the next best focal point, based on some pre-defined metric, to be presented in Section \ref{sec:active}. Following the cognitive model proposed by Itti et al. \cite{trad_sal}, as applied in modern visual models (e.g., VOCUS2 \cite{saliency}, IVSN \cite{ivsn}), we also consider an inhibition of return (IOR) mechanism that prevents the gaze from being directed to previously visited regions.

\subsection{Fusion Model for Semantic Maps} \label{sec:fusion}

In a Bayesian context, the probabilities $P \! \left( C^{\mathbf{x}} \!=\! k \mid \mathcal{I}_{1:t} \right)$, which corresponds to the posterior distributions that make up the semantic map \eqref{eq:semanticmap}, can be treated as random variables with associated uncertainty. One simple and convenient solution to characterize the underlying uncertainty consists of modeling the distributions of the map \eqref{eq:semanticmap} as Dirichlet distributions, $P \left( C^{\mathbf{x}} \!=\! k \vert \mathcal{I}_{1:t} \right) \!\sim\! \mathrm{Dir} (\boldsymbol{\beta})$, with parameters $\boldsymbol{\beta} = \left( \beta_0, \beta_1, \dots, \beta_K \right)$. We can thus represent the semantic map \eqref{eq:semanticmap} as a map of the Dirichlet $\beta$ parameter estimates at each time and location  
\begin{equation}
    \mathbf{B}_t (\mathbf{x}) = \boldsymbol{\beta}_t^{\mathbf{x}} =  \left[ \beta_{t,0}^{\mathbf{x}}, \beta_{t,1}^{\mathbf{x}}, \thinspace \dots \thinspace, \beta_{t,K}^{\mathbf{x}} \right]^T .
    \label{eq:betamap}
\end{equation}
 This allows for the inference of the posterior categorical probability distribution on different coordinates of the map, taking into account that $P \! \left( C^{\mathbf{x}}\!=\!k \thinspace \vert \thinspace \boldsymbol{\beta}^{\mathbf{x}}_{t} \right)$ is a Dirichlet-compound multinomial distribution \cite{compound}
\begin{equation}
    P \! \left( C^{\mathbf{x}}\!=\!k \thinspace \vert \thinspace \boldsymbol{\beta}^{\mathbf{x}}_{t} \right) = \dfrac{n!}{\left(\sum_{i=0}^{K} \beta_{t,i}^{\mathbf{x}}\right)^n} \prod\limits^K_{i=0} \dfrac{\left(\beta_{t,i}^{\mathbf{x}}\right)^{n_i}}{n_i !}
    \label{eq:finalsemanticmap}
\end{equation} where $n\!=\!\sum_{i=0}^{K} n_i, n_k \!=\! 1$ and $n_i \!=\! 0, \forall i \! \in\! \mathcal{C} \setminus k$.
From these parameters we can  compute the expected values and co-variances of the class distribution: 
\begin{align}
p_k &= E[C^{\mathbf{x}}=k] = \frac{\beta_k}{\sum_j \beta_j} \\
\sigma_{ij} &= Cov(C^{\mathbf{x}}=i, C^{\mathbf{x}}=j) = -p_ip_j, i\neq j
\label{eq:confscores}
\end{align}

In a visual task, we typically do not have access to the true semantic label of the observed objects, but only to the vector of scores provided by the object detector. The work developed by Kaplan et al. in \cite{kaplan} proposes some methodologies to fuse the information coming from probabilistic sensors. It establishes a comparison between Bayesian approaches (e.g. product rule, sum rule) and a proposed fusion rule (to which Dias et al. \cite{main} conveniently refer to as Kaplan rule), formulated with \eqref{eq:betamap} as 
\begin{equation}
    \beta_{t+1,k} = \dfrac{\beta_{t,k} \left( 1 + \dfrac{\lambda_{t,l,k}}{\sum_{j=0}^{K} \beta_{t,j} \lambda_{t,l,j}} \right)}{1 + \dfrac{\min_j \lambda_{t,l,j}}{\sum_{j=0}^{K} \beta_{t,j} \lambda_{t,l,j}}} .
    \label{eq:kaplanrule}
\end{equation}
where $\lambda_{t,l,k} = P(S_{t,l} \vert C=k)$ are the detector score likelihoods. Hence, assuming that our detector is perfect and that, for this reason, we do not have to account for epistemic (or model) uncertainty \cite{calibration}, we can simply consider the generated detection scores \eqref{eq:scores} as the likelihoods $P(S_{t,l} \vert C=k) = s_{t,l,k}$. Therefore, in this particular case, we consider $s_{t,l,k} = \lambda_{t,l,k}$.

Kaplan's rule \eqref{eq:kaplanrule} is a moment-matching approach \cite{kaplan} that can be applied when fusing new measurements \eqref{eq:scores} retrieved from surrounding location detections. One clear advantage of this fusion rule \cite{kaplan} is that, as a moment-matching approach, it can return a posterior Dirichlet distribution that fits the actual class posterior distribution \eqref{eq:finalsemanticmap} in terms of the first statistical moment (i.e. the mean). Essentially, the expected value of the resulting Dirichlet distribution \eqref{eq:betamap} tends to be an accurate estimate \cite{kaplan} of the actual posterior distribution's mean \eqref{eq:semanticmap}.

\subsection{Score Calibration}\label{fom}

Kaplan's rule \eqref{eq:kaplanrule} uses the detector score likelihoods instead of the actual scores (see \cite{kaplan} Eq. (1)), defined in Fig. \ref{fig:bayesnet} as $P(S \vert C\!=\!k)$. If a detector is calibrated, then $s_k = P(C=k \vert S) \propto \lambda_k$ \cite{calibration}, and the actual classifier scores can be used in Kaplan's rule \eqref{eq:kaplanrule}.
However, object classifiers are often not properly calibrated to reflect the posterior probabilities of the actual class, contrary to what is often assumed in many classification problems \cite{dirichlet}. To obtain the actual measurement likelihoods $P(S \vert C=k)$ we learn, from a large training dataset, multiple Dirichlet distributions for each class $\mathrm{Dir}(S \thinspace \vert \thinspace \boldsymbol{\alpha}_k)$, 
which can be applied in a Bayes classifier
\begin{equation}
    P\!\left( C = k \vert S \right) = \dfrac{P\! \left( S \vert C = k \right) P\!\left( C = k \right)}{P\!\left( S \right)} \propto \mathrm{Dir} \left( S \vert \boldsymbol{\alpha}_{k} \right),
    \label{eq:bayesrule}
\end{equation}
\noindent
considering $P\!\left( C = k \right)$ as a uniform prior categorical distribution, thus avoiding the introduction of bias in the model.
The Dirichlet likelihoods for each class are estimated through
\begin{equation}
    \mathrm{Dir} \! \left( S \thinspace \vert \thinspace \boldsymbol{\alpha}_{k} \right) =  \dfrac{\Gamma \! \left( \sum^K_{i=0} \alpha_{k,i} \right)}{\prod^K_{i=0} \Gamma\!\left(\alpha_{k,i}\right)} \prod\limits^K_{i=0} s_{i}^{\alpha_{k,i} - 1}
    \label{eq:dirichlet}
\end{equation}
\noindent involving the Euler's Gamma function $\Gamma(z) =  \int_0^\infty t^{z-1} e^{-t} dt$.

Essentially, each set of Dirichlet parameters is estimated by fitting multiple detection data, consisting of score vectors \eqref{eq:scores}, respectively, associated with a ground-truth class $k\in\mathcal{C}$. There is no known closed-form maximum-likelihood estimation for Dirichlet distributions. Hence, we consider a simple and efficient iterative method, proposed by Minka \cite{dirichlet}, to fit the semantic data extracted from a large dataset to the Dirichlet likelihood \eqref{eq:dirichlet}. This culminates in several sets of $\alpha$ parameters
\begin{equation}
    \boldsymbol{\alpha}_{k} = \left( \alpha_{k,0}, \alpha_{k,1}, \thinspace \dots \thinspace , \alpha_{k,K} \right)
    \label{eq:alphas_fom}
\end{equation}
\noindent with which we capture the aleatoric uncertainty, derived from multiple data-related factors, such as illumination or occlusion. 

\subsection{Foveal Observation Model}\label{sec:fom}

\begin{figure}
\centering
\subfigure{\label{fig:original}\includegraphics[width=0.49\columnwidth]{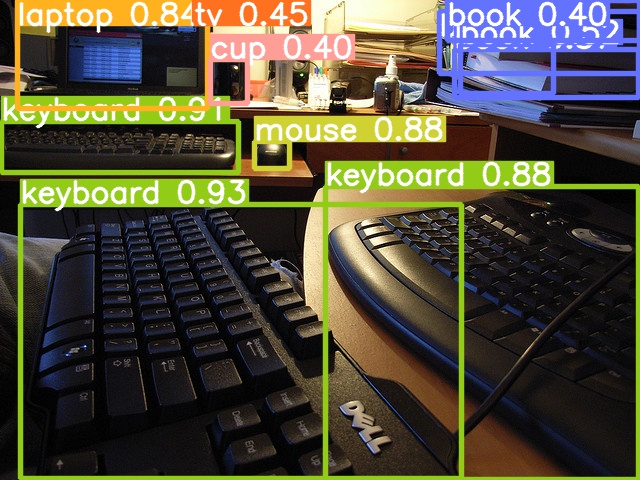}} 
\subfigure{\label{fig:ambiguous}\includegraphics[width=0.49\columnwidth]{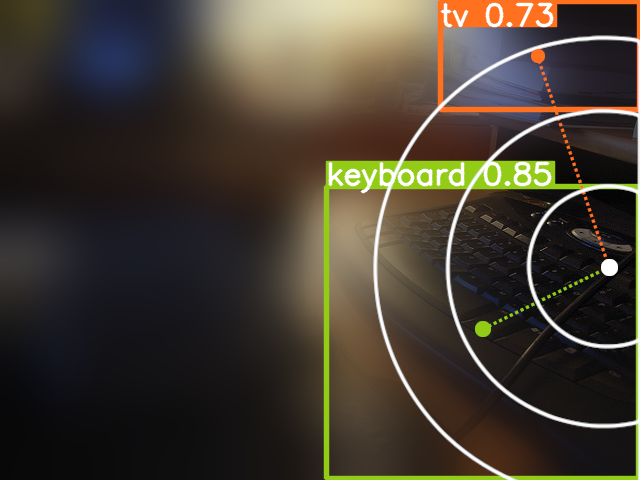}}  
\caption{Example of an ambiguous detection due to peripheral distortion \cite{foveal}. The semantic content located on the top-right corner of the simulated visual field corresponds to the class \textit{'book'} (left). Nonetheless, the class predicted by the detector is \textit{'tv'} (right), due to the influence of the foveal sensor's position.}
\label{fig:ambiguities}
\end{figure}

The incorporation of a foveal visual system generates extra data-induced uncertainty in the outputs of the object detection model \cite{activevision}, as shown in Fig. \ref{fig:ambiguities}. Taking advantage of the radial distribution of blur across the visual field, a foveal observation model \cite{main} is designed to calibrate the scores \eqref{eq:scores}, directly outputted by the detector. However, the calibration model now depends not only on the object class but also on the distance to the fovea, as the classifier scores are significantly affected by the different levels of blur across the simulated visual field.

\noindent This calibration technique considers the relative distance from the center of the prediction's bounding-box $\mathcal{B}_{t,l}$ to the center of the fovea, $d = \| (u,v) \|$, as illustrated in Fig. \ref{fig:map}. The pair $(u,v)$ represents the set of central coordinates of $\mathcal{B}_{t,l}$ relative to the center of the fovea. The foveal observation model, introduced in Fig. \ref{fig:general} as the method used for uncertainty characterization and represented in Fig. \ref{fig:bayesnet} by a plate, essentially comprises multiple sets of Dirichlet parameters for each class and distance to the fovea (in a discrete set), specifically a structure that contains $(K\!+\!1) \times N$ arrays of $\boldsymbol{\alpha}_{k,d}$ parameters. This means that there is a set of parameters \eqref{eq:alphas_fom} for each class in all $N$ distance levels. 
These levels consist of discretizations of the focal distance $(d)$ that grant the existence of a finite number of compartmentalized Dirichlet distributions that are trained to model the $p\left( S \vert C = k, d \right)$ likelihood distribution. From this likelihood, we compute $p\left( C = k \vert S, d \right)$, with a small adaptation of the Bayes Rule \eqref{eq:bayesrule}. In this adaptation, we consider the $\mathrm{Dir} \! \left( S \thinspace \vert \thinspace \boldsymbol{\alpha}_{k,d} \right)$ distribution as the estimated likelihood, $p\left( S \vert C = k, d \right)$, with which we can model the $p\left( C = k \vert S, d \right)$ posterior distribution. 
Hence, a new set of calibrated scores $S^\prime_{t,l}$ proportional to the measurement likelihoods can be obtained through \eqref{eq:bayesrule}
\begin{align}
\begin{split}
    S^\prime &= \left( s^\prime_{0}, s^\prime_{1}, \thinspace \ldots \thinspace,  s^\prime_{K} \right) \\
    &= \dfrac{1}{D}\left[ p\!\left( S \vert C = 0 , d \right) , \thinspace \ldots \thinspace, p\!\left( S \vert C = K , d\right) \right]^T\\
    &= \dfrac{1}{D} \left[ \mathrm{Dir} \! \left( S \thinspace \vert \thinspace \boldsymbol{\alpha}_{0,d} \right) , \thinspace \ldots \thinspace, \mathrm{Dir} \! \left( S \thinspace \vert \thinspace \boldsymbol{\alpha}_{K,d} \right) \right]^T
    \label{eq:corrected_scores}
\end{split}
\end{align}
\noindent where $D \!=\! \sum^K_{k=0} \mathrm{Dir} \! \left( S \thinspace \vert \thinspace \boldsymbol{\alpha}_{k,d} \right)$ represents the normalization factor which grants that $\sum^K_{k=0} s^\prime_{k} = 1$. Notice that, since Kaplan's rule \eqref{eq:kaplanrule} uses directly the likelihoods instead of the posterior probabilities, the normalization of the likelihoods is not ultimately relevant for this process, and is hereby defined only to demonstrate that the calibrated scores \eqref{eq:corrected_scores} can be conformed into a categorical distribution.
The prior distribution $p\!\left( C = k \mid d \right)$, which consequently appears in the adaptation of \eqref{eq:bayesrule}, is assumed to be uniform, as an object can be arbitrarily located within the confines of a scene. Kaplan's classifier update rule \eqref{eq:kaplanrule} can also accommodate foveal calibrated scores since we can directly replace raw scores \eqref{eq:scores} with corrected foveal scores \eqref{eq:corrected_scores} in \eqref{eq:kaplanrule}. This is identified in Fig. \ref{fig:general} as the calibrated semantic map, which can be found along the dotted path, right after the uncertainty characterization block.

\subsection{Active Perception Model}\label{sec:active}

Let us now turn our focus to the active perception model, which operates based on the information collected on the semantic map \eqref{eq:betamap}. At a given timestamp $t$, the current state of \eqref{eq:betamap} is consulted by the active perception mechanism, to determine the most promising cell $(x^\ast_{t+1},y^\ast_{t+1})$ where the gaze is to be shifted at the $t+1$ instant, according to a pre-defined task-specific metric.

\subsubsection{Non-Predictive Visual Search}

Taking into account the dominant greedy nature of the visual search task \cite{visualsearch}, the priority is to find the region where the target object is localized and to orient the gaze toward it. Therefore, we start by considering a simple non-predictive optimization problem, consisting of maximizing the posterior \eqref{eq:semanticmap} of a given target class $k \in \mathcal{C}$, as 
\begin{equation}
    \left( x^\ast_{t+1},y^\ast_{t+1} \right) = \mathbf{x}^\ast = \argmax\limits_{\mathbf{x}} \left\{ \dfrac{\beta^{\mathbf{x}}_{t,k}}{\sum^K_{j=0} \beta^{\mathbf{x}}_{t,j}} \right\} .
    \label{eq:peaks_nonpred}
\end{equation}
This intuitive approach consists of consulting the current state of the map \eqref{eq:betamap} in order to infer the class confidence in each pair of coordinates $(x,y)$ in the context grid. Then, similarly to traditional cognitive approaches \cite{saliency,trad_sal}, we use a winner-take-all 
(WTA) mechanism that selects the next best focal point.

 \subsubsection{Predictive Visual Search}

Although it is not the first attempt to combine object detection and active perception to perform visual search \cite{objsearch}, this is the first work to integrate the semantic-based active perception model, introduced by Dias \cite{main}, into a visual target search task with foveal vision. Therefore, we wish to evaluate the full potential of the active perception approach, by also including an update simulation, in order to predict the posterior probabilities \eqref{eq:semanticmap}. Taking this objective into account and following the proposed methodology \cite{main}, we simulate a complete map update \eqref{eq:betamap} for each possible focal cell, depending on $\mathbf{f}^\prime\!=\!(x^\prime,y^\prime)$, where the fovea could be next centered. We define the predicted semantic maps as
\begin{equation}
   \Bar{\mathbf{B}}^{\mathbf{f}^\prime}_{t+1} (\mathbf{x}) = \mathbb{E} \left[ \mathbf{B}_{t+1} (\mathbf{x}) \thinspace \vert \thinspace \mathbf{B}_t (\mathbf{x}), \mathbf{f}^\prime \right],
    \label{eq:predbeta}
\end{equation}
\noindent which simulates the expected value of the semantic map after a fixation to coordinate $\mathbf{f}^\prime$ is executed. 
This means that the active perception mechanism has to select one of $X \times Y$ possible scenarios, based on the knowledge that results from each simulation.
To solve \eqref{eq:predbeta}, we first compute the expected value of the scores that should be outputted by the object detector for the coordinates $\mathbf{x}$, if the focal point is set in $\mathbf{f}^\prime$, through the expansion of the posterior distribution
\begin{align}
    \begin{split}
    &p \left( S^{\mathbf{x}} \thinspace \vert \thinspace \mathbf{f}^\prime, \boldsymbol{\beta}^{\mathbf{x}}_{t} \right) =\\ &= \sum\limits^K_{k=0} p \left( S^{\mathbf{x}}, C^{\mathbf{x}}\!=\!k \thinspace \vert \thinspace \mathbf{f}^\prime, \boldsymbol{\beta}^{\mathbf{x}}_{t} \right) \\ &= \sum\limits^K_{k=0} p \left( S^{\mathbf{x}} \thinspace \vert \thinspace C^{\mathbf{x}}\!\!=\!k, \mathbf{f}^\prime, \boldsymbol{\beta}^{\mathbf{x}}_{t} \right) P \! \left( C^{\mathbf{x}}\!\!=\!k \thinspace \vert \thinspace \mathbf{f}^\prime, \boldsymbol{\beta}^{\mathbf{x}}_{t} \right) \\ &= \sum\limits^K_{k=0} p \left( S^{\mathbf{x}} \thinspace \vert \thinspace C^{\mathbf{x}}\!=\!k, \mathbf{f}^\prime \right) P \! \left( C^{\mathbf{x}}\!=\!k \thinspace \vert \thinspace \boldsymbol{\beta}^{\mathbf{x}}_{t} \right).
    \end{split}
    \label{eq:probpred}
\end{align}

The Dirichlet-compound multinomial \cite{compound} posterior class probabilities $ P \left(C^{\mathbf{x}} = k \mid \boldsymbol{\beta}^{\mathbf{x}}_{t} \right) = \beta_{t,k}^{\mathbf{x}} \mathbin{/} \sum_{j=0}^{K} \beta_{t,j}^{\mathbf{x}}$ can be inferred through \eqref{eq:finalsemanticmap}.
The steps presented in \eqref{eq:probpred} take advantage of the conditional independence between $S^{\mathbf{x}}$ and $\boldsymbol{\beta}_t^{\mathbf{x}}$ for a given class $k$. It also depends on the fact that the class to which the object contained in a $\mathbf{x} = (x,y)$ cell belongs is not constrained by the new location of the center of the fovea $\mathbf{f}^\prime = (x^\prime,y^\prime)$. Taking advantage of the linearity of the expectation operator applied to \eqref{eq:probpred}, we can then assert that
\begin{equation}
    \mathbb{E} \left[ S^{\mathbf{x}} \thinspace \vert \thinspace \mathbf{f}^\prime, \boldsymbol{\beta}^{\mathbf{x}}_{t} \right] = \sum\limits^K_{k=0} \mathbb{E} \left[ S^{\mathbf{x}} \thinspace \vert \thinspace C^{\mathbf{x}} \! = \! k, \mathbf{f}^\prime \right] P\!\left(C^{\mathbf{x}} \! = \!k \thinspace \vert \thinspace \boldsymbol{\beta}^{\mathbf{x}}_{t} \right)
    \label{eq:proportional}
\end{equation}
\noindent meaning that the computation of predictive scores is hanging on the determination of $\mathbb{E} \left[ S^{\mathbf{x}} \mid C^{\mathbf{x}} \!=\! k, \mathbf{f}^\prime \right]$ in each iteration.
Consider now that $d^\prime = \| \mathbf{x} - \mathbf{f}^\prime \| = \| (u^\prime,v^\prime) \|$ represents the distance level associated with a given $\mathbf{x}$ cell in relation to the central coordinates of the $\mathbf{f}^\prime$ cell, where the gaze will be shifted, measured with the $(u^\prime,v^\prime) = (x - x^\prime, y - y^\prime)$ relative coordinates, where $(x,y)$ and $(x^\prime,y^\prime)$ are the central pixel coordinates of the referred $\mathbf{x}$ and $\mathbf{f^\prime}$ cells, respectively. Hence, the expected score vectors can be computed as
\begin{equation}
    \Bar{S}^{\mathbf{x}} = \mathbb{E} \! \left[ S^{\mathbf{x}} \thinspace \vert \thinspace \mathbf{f}^\prime, \boldsymbol{\beta}^{\mathbf{x}}_{t} \right]  = \sum\limits^K_{k=0} \dfrac{\boldsymbol{\alpha}_{k,d^\prime}}{\sum^K_{j=0} \alpha_{k,d^\prime\!,j}} \dfrac{\beta^{\mathbf{x}}_{t,k}}{\sum^K_{j=0} \beta^{\mathbf{x}}_{t,j}}
    \label{eq:expected_operator}
\end{equation}
\noindent which can then be applied with Kaplan's rule \eqref{eq:kaplanrule} to update the predicted semantic map \eqref{eq:predbeta}. From then on, we can infer the estimated posterior distributions \eqref{eq:semanticmap} out of $\Bar{\mathbf{B}}^{\mathbf{f}^\prime}_{t+1}$ \eqref{eq:predbeta}, through \eqref{eq:finalsemanticmap}, determining the cell most promising, to where the gaze is to be shifted, based on a given task-dependent metric \cite{apfov}.

For the predictive visual search task, given its greedy nature \cite{visualsearch}, we will not need complete updates of the predicted semantic maps. In fact, only the scores of the detections in the high-resolution area close to the next fixation point $\mathbf{f}^\prime$ are relevant. In other words, we assume that the best fixation point will be close to the object of interest. Thus, for each possible fixation point $\mathbf{f}^\prime$, we only need to predict the detector scores at the foveal locations ($d=0$)
\begin{equation}
    \Bar{S}^{\mathbf{x}}_0 = \sum\limits^K_{k=0} \dfrac{\boldsymbol{\alpha}_{k,0}}{\sum^K_{j=0} \alpha_{k,0,j}} \dfrac{\beta^{\mathbf{x}}_{t,k}}{\sum^K_{j=0} \beta^{\mathbf{x}}_{t,j}},
    \label{eq:local_scores}
\end{equation}
\noindent using only the set of Dirichlet distributions, with parameters \eqref{eq:alphas_fom}, that correspond to the innermost distance level.

Similarly to the non-predictive approach \eqref{eq:peaks_nonpred}, we use a WTA mechanism that selects the next best focal point through
\begin{equation}
    \left( x^\ast_{t+1},y^\ast_{t+1} \right) = \mathbf{x}^\ast = \argmax\limits_{\mathbf{x}} \left\{ \dfrac{\Bar{\beta}^{\mathbf{x}}_{t+1,C}}{\sum^K_{k=0} \Bar{\beta}^{\mathbf{x}}_{t+1,k}} \right\} ,
    \label{eq:peaks_pred}
\end{equation}
\noindent where $\Bar{\beta}^{\mathbf{x}}_{t+1,C}$ represents the expected Dirichlet parameter for class $C$, extracted from the local $\mathbf{f}^\prime\! = \mathbf{x}$ predictive map $\Bar{\mathbf{B}}^{\mathbf{f}^\prime\! = \mathbf{x}}_{t+1}$. 

\subsubsection{Scene Exploration}\label{sec:results_sceneexp}


In the exploration task, the goal is to minimize the overall uncertainty in the map while performing a minimal number of actions (i.e. gaze shifts). The gaze shift will be made to the new location of the fixation that minimizes the uncertainty of the predicted map \eqref{eq:predbeta}. One form of quantifying uncertainty consists of computing the distance between the semantic map distributions and non-informative distributions \cite{main}. We consider the Kullback-Leibler divergence \cite{bishop} as a measurement of the distance between the estimated Dirichlet distribution and a base distribution, as
\begin{equation}
    F^{\mathbf{x}}_{t} = \mathcal{D}_{KL} \left( \mathrm{Dir} \left( \boldsymbol{\beta}_t^{\mathbf{x}} \right) \vert \vert \thinspace \mathrm{Dir} \left( \boldsymbol{\beta}^0 \right) \right) ,
    \label{eq:kldivergence}
\end{equation}
\noindent where $\boldsymbol{\beta}^0$ corresponds to the initial uniform Dirichlet distribution $(\beta^0_k = 0.5 \thinspace, \forall k \! \in \! \{ 0, \dots, K\})$, representing a state of maximum uncertainty, that is initially warranted to each $\mathbf{x}$ cell. Intuitively, we want to select the cell that maximizes the distance between the two Dirichlet distributions \cite{main} with \eqref{eq:kldivergence}. 
The classification negentropy \cite{bishop} is another uncertainty metric considered for active perception, formulated as
\begin{equation}
    F^{\mathbf{x}}_{t} = \sum\limits_{k=0}^K \dfrac{\beta^{\mathbf{x}}_{t,k}}{\sum^K_{j=0} \beta^{\mathbf{x}}_{t,j}} \log \left( \dfrac{\beta^{\mathbf{x}}_{t,k}}{\sum^K_{j=0} \beta^{\mathbf{x}}_{t,j}} \right) .
    \label{eq:entropy}
\end{equation}
\noindent The higher the negentropy, the lower the cell's confusion. For this reason, we consider applying both the Kullback-Leibler divergence \eqref{eq:kldivergence} and the negentropy \eqref{eq:entropy} metrics with a simple acquisition function that transverses each map \eqref{eq:predbeta}, accumulating the expected results of the selected metric. With this approach, we select the location $\mathbf{f}^\prime = (x^\prime,y^\prime)$ that maximizes the sum of the expected metric results over \eqref{eq:probpred} as 
\begin{equation}
    \left( x^\ast_{t+1},y^\ast_{t+1} \right) = \mathbf{x}^\ast = \argmax\limits_{\mathbf{f}^\prime} \left\{ \sum\limits_{\mathbf{x}} \mathbb{E} \left[ F^{\mathbf{x}}_{t+1} \thinspace \vert \thinspace \mathbf{f}^\prime \right] \right\} ,
    \label{eq:minimize}
\end{equation}
\noindent where $\mathbb{E} \!\left[ F^{\mathbf{x}}_{t+1} \mid \mathbf{f}^\prime \right]$ represents the expected measurement of the metric $F$ for the next iteration (i.e. at the $t+1$ timestamp), assuming that the focal point is then set on the $\mathbf{f}^\prime$ coordinates.
Finally, the last alternative metric consists of computing the absolute difference between the categorical scores of the two most probable classes \eqref{eq:semanticmap}, which can be expressed with \eqref{eq:betamap} as
\begin{equation}
    F^{\mathbf{x}}_{t} = \max\limits_k \left\{ \dfrac{\beta^{\mathbf{x}}_{t,k}}{\sum^K_{j=0} \beta^{\mathbf{x}}_{t,j}} \right\} - \max\limits_{k \setminus \varkappa} \left\{ \dfrac{\beta^{\mathbf{x}}_{t,k}}{\sum^K_{j=0} \beta^{\mathbf{x}}_{t,j}} \right\},
    \label{eq:twopeaks}
\end{equation}
\noindent where $\varkappa = \argmax\limits_k \left\{ \beta_{t,k}^{\mathbf{x}} \mathbin{/} \sum_{j=0}^{K} \beta_{t,j}^{\mathbf{x}} \right\}$ represents the most probable class, taking advantage of the knowledge in \eqref{eq:betamap}. Following the work of Dias et al. \cite{main}, we consider applying this metric with a different acquisition function \cite{apfov}, known as the expected improvement.  This function aims at maximizing the expected magnitude of the two-peaks improvement \eqref{eq:twopeaks} as
\begin{equation}
    \hspace{-0.1cm}
    \left( x^\ast_{t+1},y^\ast_{t+1} \right) = \mathbf{x}^\ast \!= \argmax\limits_{\mathbf{f}^\prime} \! \left\{ \!\max\limits_{\mathbf{x}} \left| \mathbb{E} \!\left[ F^{\mathbf{x}}_{t+1} \vert \thinspace \mathbf{f}^\prime \right] \!-\! F^{\mathbf{x}}_t \right| \right\}
    \label{eq:maximize}
\end{equation}
\noindent which is a greedy approach, since it consists of selecting the maximum absolute difference, out of all $\mathbf{x}$ map coordinates, for each $\mathbf{f}^\prime$, that represents all the possible next focal points.

\section{Experiments \& Results}\label{sec:exp}

In this section, we intend to highlight some important aspects of our testing setup and present the most relevant experimental results, consisting of the evaluation of cognitive attention models, proposed and presented in Section \ref{sec:methods}, during the completion of active visual perception experiments involving scene exploration and visual search tasks.

\subsection{Overview of the Experimental Setup}

\begin{figure}
\centering
\includegraphics[width=1\columnwidth]{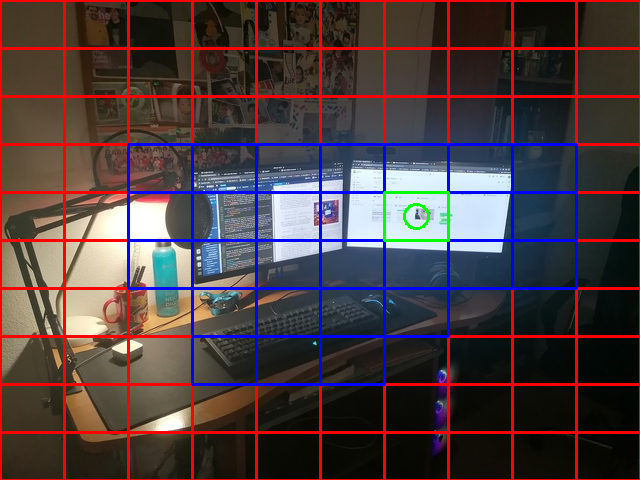}
\caption{Grid of (red) cells dividing the semantic map of Fig. \ref{fig:detection_example}. The green-colored cell that is visible on the 10x10 grid marks the focal point and the cells that are updated with the YOLOv3 detections are highlighted in blue.}
\label{fig:grid_example}
\end{figure}

First, to generate predictions, as depicted in Fig. \ref{fig:map}, we select YOLOv3 \cite{yolo}, which is a deep detection model that follows the one-step framework \cite{odreview}, based on global regression/classification. Although not the latest version, YOLOv3 remains a fast and accurate model for object detection that uses Darknet-53 \cite{yolo} as its CNN backbone architecture.

The foveal observation model is trained with the full COCO 2017 training set \cite{coco}, foveated around random focal points, using Minka's algorithm \cite{dirichlet} to fit the categorical scores \eqref{eq:scores} generated by YOLOv3, into multiple Dirichlet distributions \eqref{eq:alphas_fom}, according to the corresponding ground truth object and focal distance level. Our validation dataset comprises 300 quasi-randomly selected images from the COCO validation set.
Essentially, each selected image must contain at least a target object and multiple distractors, adding up to a grand total of 8 or more different instances. Furthermore, any target object contained in a selected image can only occupy and be associated with, at most, 20\% of the total area of the fixed field of view, to promote a fair iterative search process and assess the influence of peripheral vision.
The image is divided into a grid of 10$\times$10 to implement the semantic map cells, as depicted in Fig. \ref{fig:grid_example}, which delimits the regions to which the gaze can be shifted. 
When initializing an experiment, each cell of the semantic map \eqref{eq:betamap} is defined by a uniform Dirichlet distribution $\boldsymbol{\beta}^0$, where $\beta^0_k = 0.5,  \forall k \in \{0, \dots, K\}$, representing an initial non-informative state. The initial focal point is also pseudo-randomly selected, as the only constraint being applied consists of not selecting a starting cell containing an instance of the targeted class. Suppose that multiple detections overlap a cell with $(x,y)$ coordinates. In that case, Kaplan's rule \eqref{eq:kaplanrule} is applied sequentially, in no particular order, with every single score vector, with \eqref{eq:corrected_scores} or without \eqref{eq:scores} foveal calibration.

\begin{figure}
\centering
\includegraphics[width=1\columnwidth]{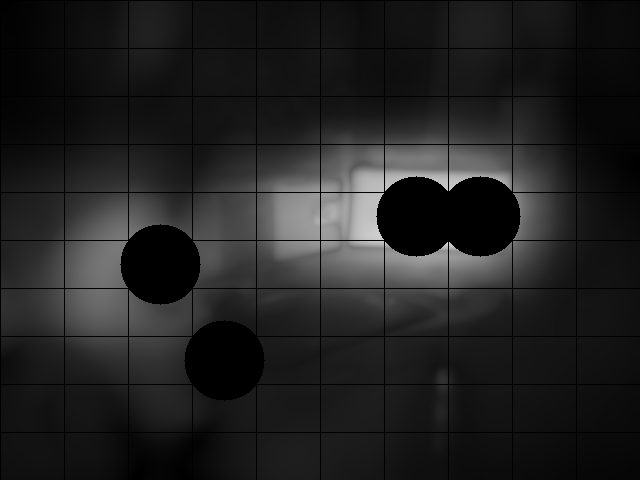}
\caption{The saliency map generated by VOCUS2 \cite{saliency} for the same foveal scene, after 4 algorithm iterations (applying the IOR mechanism), is also presented.}
\label{fig:saliency_example}
\end{figure}

Regarding scene exploration, 
our aim is to conclude whether the active perception model is able to reduce the confusion inherent to the semantic map \eqref{eq:betamap} with the minimal amount of gaze fizations. For this reason, we assess the performance of the methods, presented in Section \ref{sec:active}, with the success rate metric, which consists of the ratio between the number of map cells where the highest categorical score \eqref{eq:finalsemanticmap} fits the class of an overlapping ground-truth object (true positives), and the total number of cells containing ground-truth objects (true positives + false negatives).

Regarding the visual search task, we figure out whether it is possible to navigate a visual scene, seeking an instance of a pre-defined class, when applying the approach described in Fig. \ref{fig:general}. Despite the broad acceptance criteria for success in visual search, related to the number of target cells, it is crucial to understand to what extent the semantic information model is able to leverage the accuracy with which the task is completed. Following the setup presented in \cite{ivsn}, a \textit{ Oracle} is responsible for terminating the search algorithm once the focal point is set on a target cell, indicating that the search has been concluded because one instance of the target class has been found. We evaluate the success in each iteration as the ratio between the number of images in which the target object has already been located at that iteration and the total number of images in the validation dataset, i.e. the cumulative performance \cite{ivsn}.

\subsection{Baseline Saliency-Based Model}

To attempt a fair comparison with an attention-based model that imitates biological cognitive mechanisms, we consider a classical conspicuity model \cite{trad_sal} that is adapted to govern perceptive decisions (dashed line in Fig. \ref{fig:general}). This constitutes what is commonly called a free-viewing model \cite{predvisualfix} as the decision process is not influenced by any task-related intricacies.

\noindent For this purpose, we implement VOCUS2 \cite{saliency}, a reformed version of Itti's traditional saliency model \cite{trad_sal}, to extract low-level features inherent to pixel-level color contrasts, in compliance with the FIT. Despite lacking top-down features (considered in recent visual attention models \cite{ivsn}), which are essential for approximating human neurological models to their full potential, VOCUS2 is still able to simulate critical aspects that guide the mechanisms that govern human perception.
In Fig. \ref{fig:saliency_example} we present a visualization of the IOR mechanism, applied on the saliency map produced by VOCUS2 after each iteration, nullifying the intensity of the pixels contained inside the previously visited regions (confined in a grid of cells \cite{main}).

\subsection{Visual Search}

In the search experiments, we assess how many gaze shifts are necessary to achieve success, with each experiment running for a total of 30 iterations. For each image from the validation dataset, we repeat the experiment 10 times, starting from different initial focal points (i.e. cells) to test whether the model is exposed to different information in every repetition. Success is achieved when one of the evaluated models leads the gaze to be shifted toward a cell that contains an instance of the class that is being targeted \cite{visualsearch} in each experiment. 

\subsubsection{Role of Prediction and Calibration}

\begin{figure}
\centering
\includegraphics[width=\columnwidth]{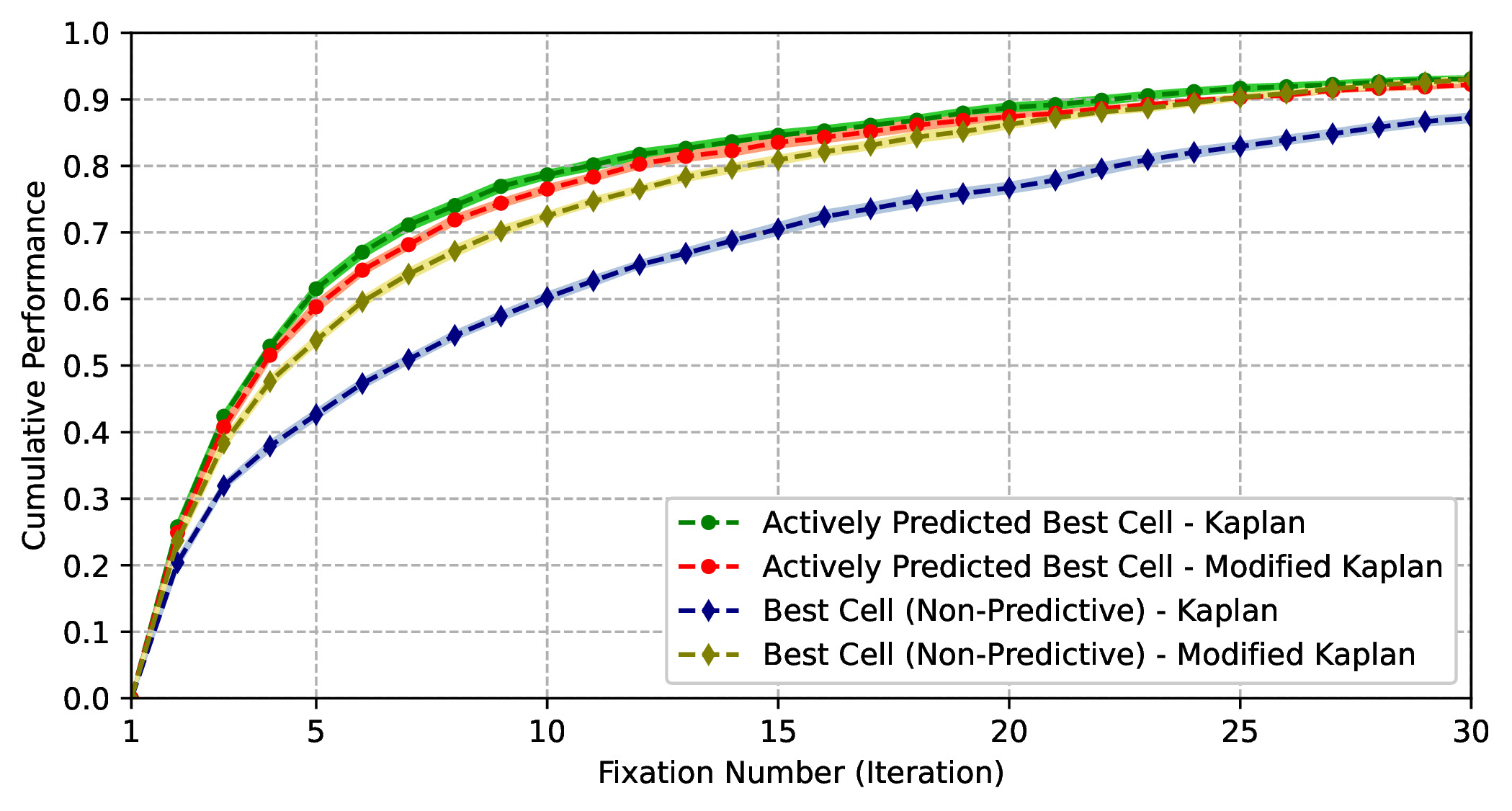}
\caption{Comparison between the mean values of the cumulative performance and the respective standard error of the mean (SEM) bands from results observed over 10 experiments, obtained using predictive and non-predictive approaches that operate on the semantic information updated either with raw (Kaplan) or calibrated scores (Modified Kaplan).}
\label{fig:cum1}
\end{figure}
We compare the two proposed methods of active visual search: non-predictive active perception \eqref{eq:peaks_nonpred}, using only the current information gathered and stored on the semantic map \eqref{eq:betamap}, and predictive active perception \eqref{eq:peaks_pred}, with simulated updates.
Furthermore, we investigate potential benefits that may proceed from implementing the foveal correction model \eqref{eq:corrected_scores}, determining whether the calibrated map can facilitate the efficient identification of cells that contain instances of the target class. Two main conclusions arise from the results that are portrayed in Fig. \ref{fig:cum1}. The first conclusion emerges from the comparison between the non-predictive search algorithms.
From the results highlighted in Tab. \ref{tab1} we can conclude that, using the corrected scores \eqref{eq:corrected_scores}, calibrated by the foveal observation model \ref{sec:fom}, appears to be considerably beneficial for the non-predictive approach, but not influencing the predictive approach. Foveal calibration leads the non-predictive methods to a performance level that approximates the results obtained with the predictive methods while consuming much fewer computational resources.
\begin{table}
\caption{Comparison of Predictive and Non-Predictive Approaches}
\begin{center}
\begin{tabular}{|c|c|c|c|c|}
\hline
\textbf{Evaluation}&\multicolumn{2}{|c|}{\textbf{Predictive}}&\multicolumn{2}{|c|}{\textbf{Non-Predictive}}  \\
\cline{2-5} 
\textbf{Metric} & \textbf{\textit{Non-Calib}} & \textbf{\textit{Calib}}& \textbf{Non-Calib}& \textbf{\textit{Calib}} \\
\hline
 CP$^{\mathrm{1}}$ @ 5 $\thinspace$  & 61,53\% & 58,83\% & 42,63\% & 53,80\% \\
 CP$^{\mathrm{1}}$ @ 15              & 84,60\% & 83,53\% & 70,53\% & 80,90\% \\
 CP$^{\mathrm{1}}$ @ 30              & 93,10\% & 92,23\% & 87,23\% & 93,03\% \\
\hline
 Max. SEM$^{\mathrm{2}}$   & 0,772\% & 0,820\% & 0,845\% & 0,718\% \\
 Comp. Cost$^{\mathrm{3}}$ & 77,81 ms& 83,76 ms& 0,144 ms& 0,133 ms\\
\hline
\multicolumn{5}{l}{$^{\mathrm{1}}$Cumulative Performance. $^{\mathrm{2}}$Standard Error of the Mean. $^{\mathrm{3}}$Per Iteration.}
\end{tabular}
\label{tab1}
\end{center}
\end{table}  
\begin{figure*}
\centering \subfigure{\label{fig:field1}\includegraphics[width=0.24\linewidth]{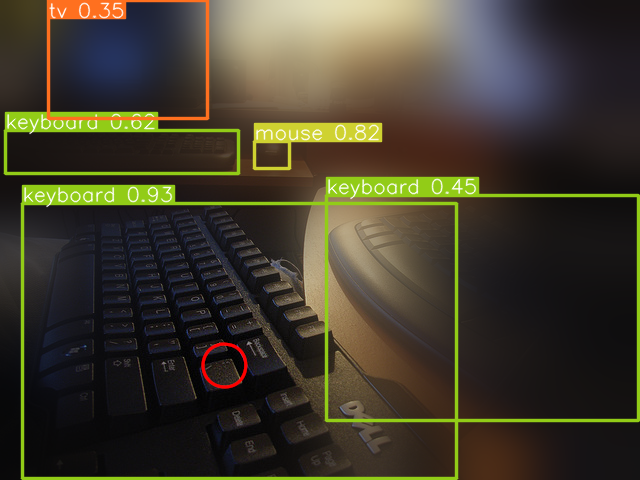}} \subfigure{\label{fig:field2}\includegraphics[width=0.24\linewidth]{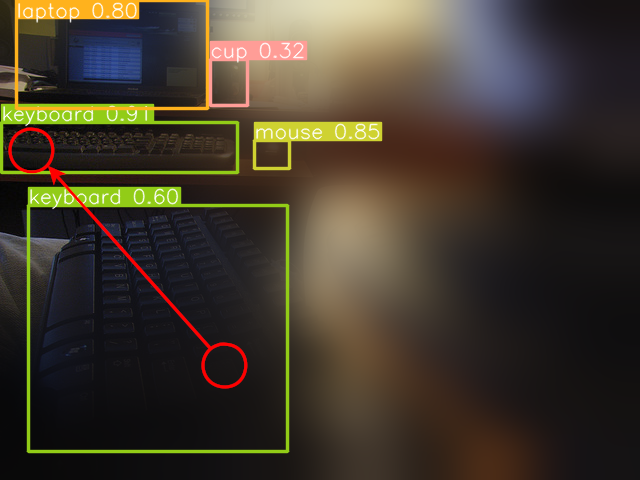}}  \subfigure{\label{fig:field3}\includegraphics[width=0.24\linewidth]{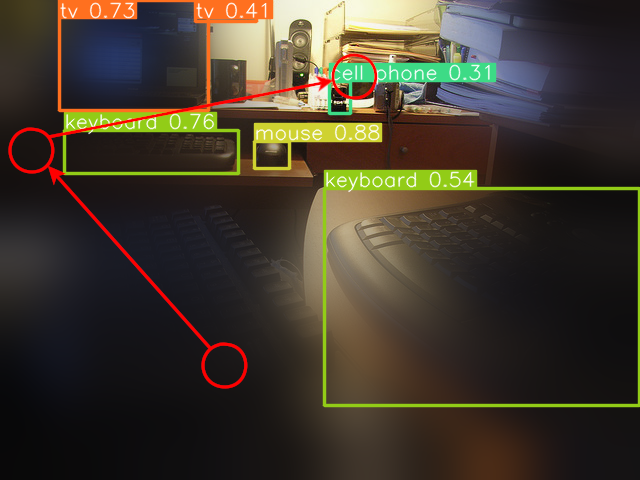}}    \subfigure{\label{fig:field4}\includegraphics[width=0.24\linewidth]{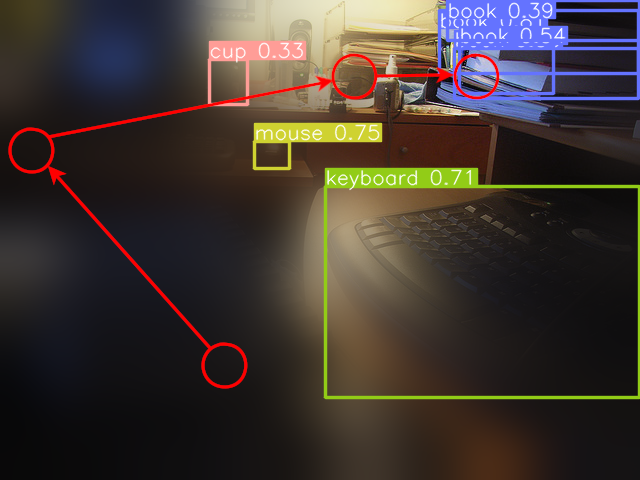}}
\subfigure{\label{fig:bars1}\includegraphics[width=0.24\linewidth]{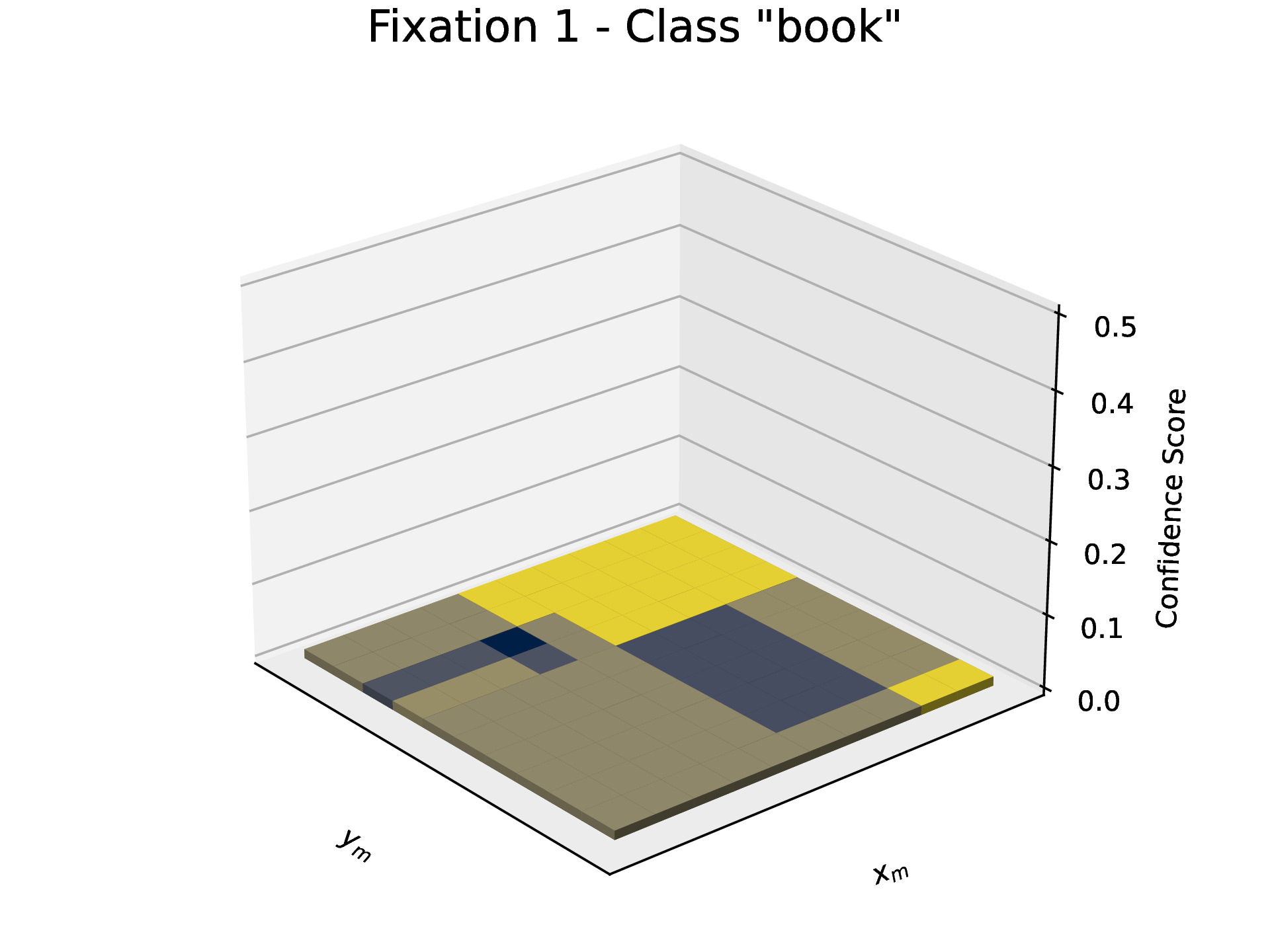}} \subfigure{\label{fig:bars2}\includegraphics[width=0.24\linewidth]{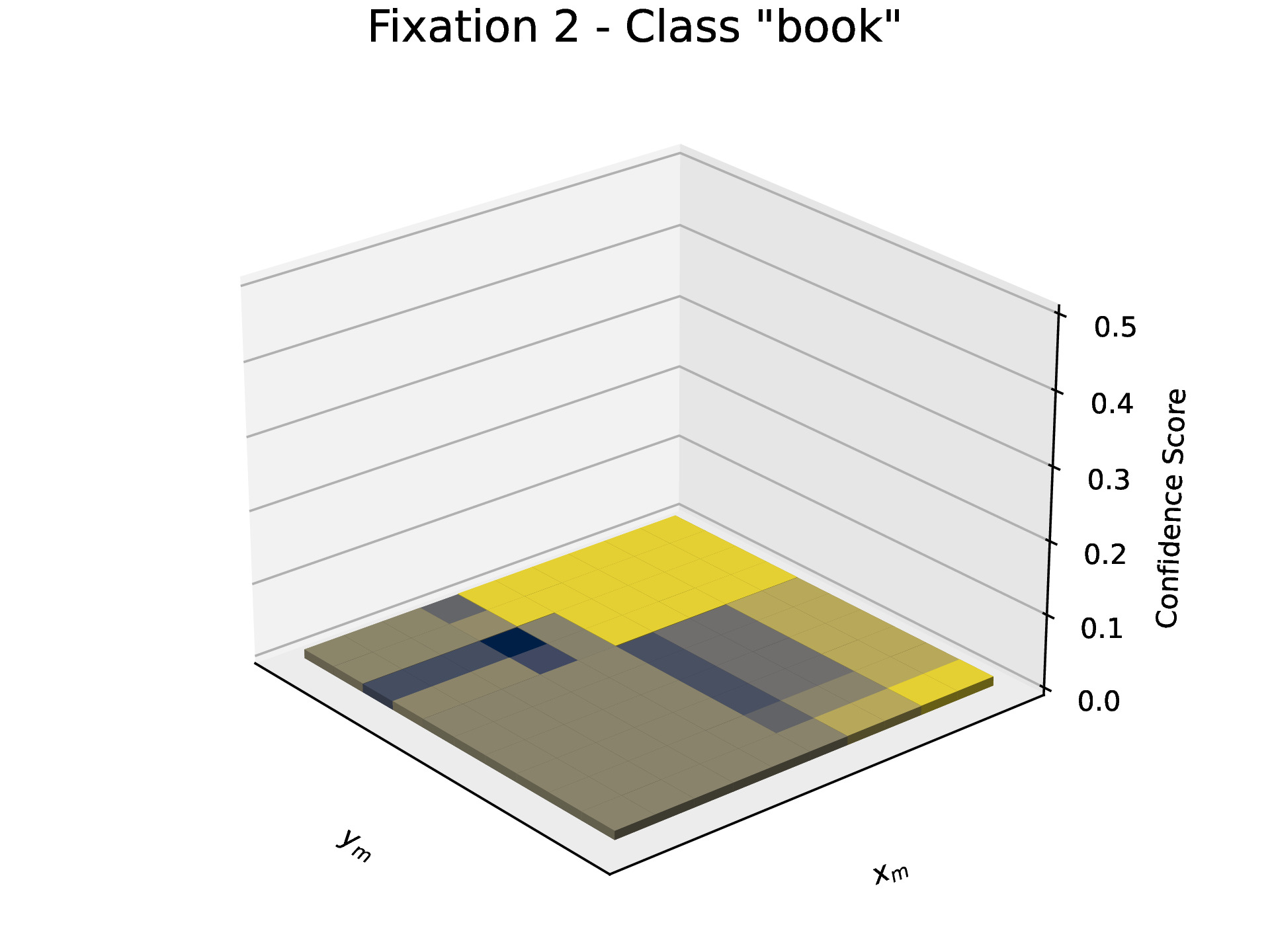}}  \subfigure{\label{fig:bars3}\includegraphics[width=0.24\linewidth]{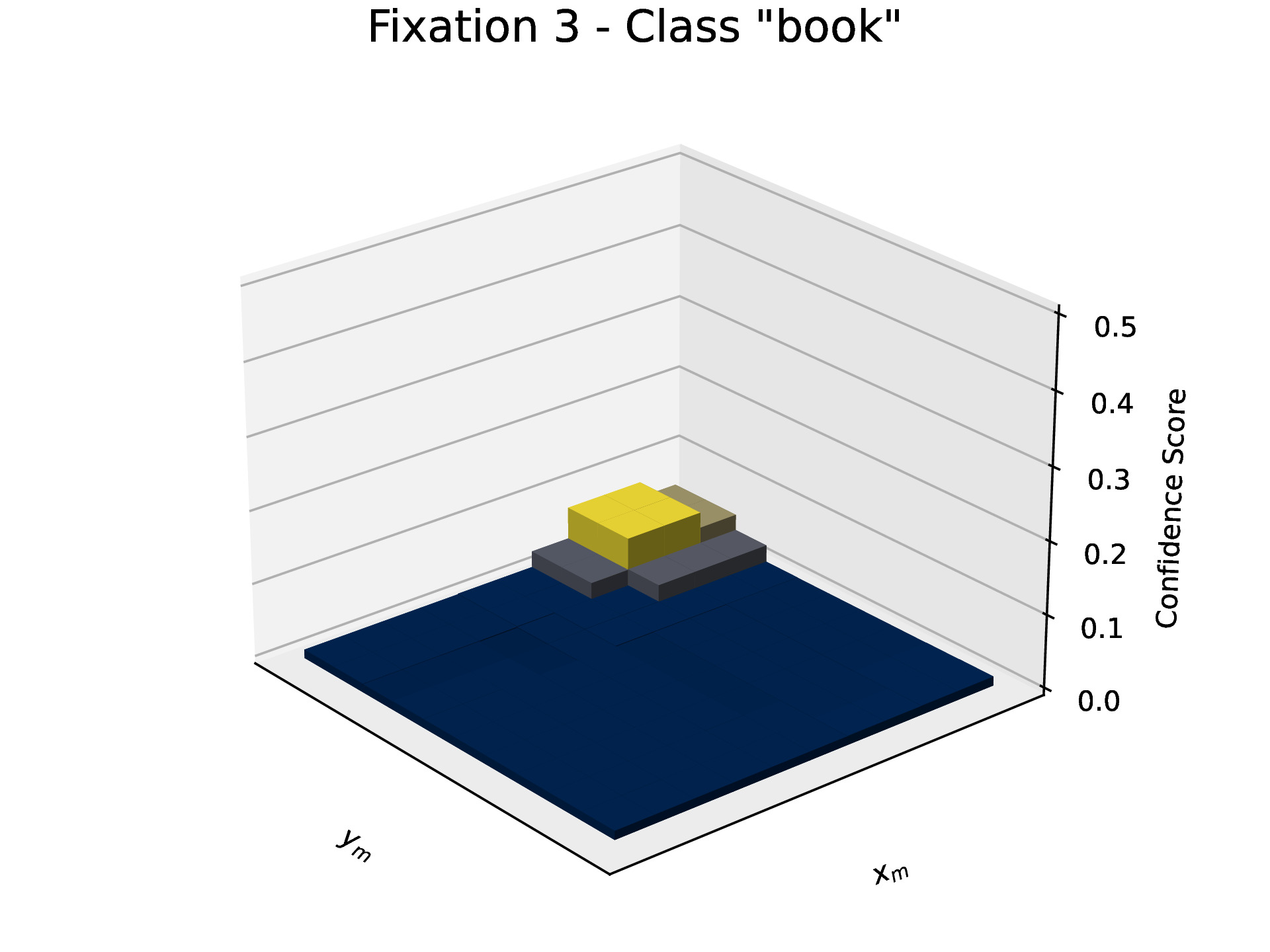}}    \subfigure{\label{fig:bars4}\includegraphics[width=0.24\linewidth]{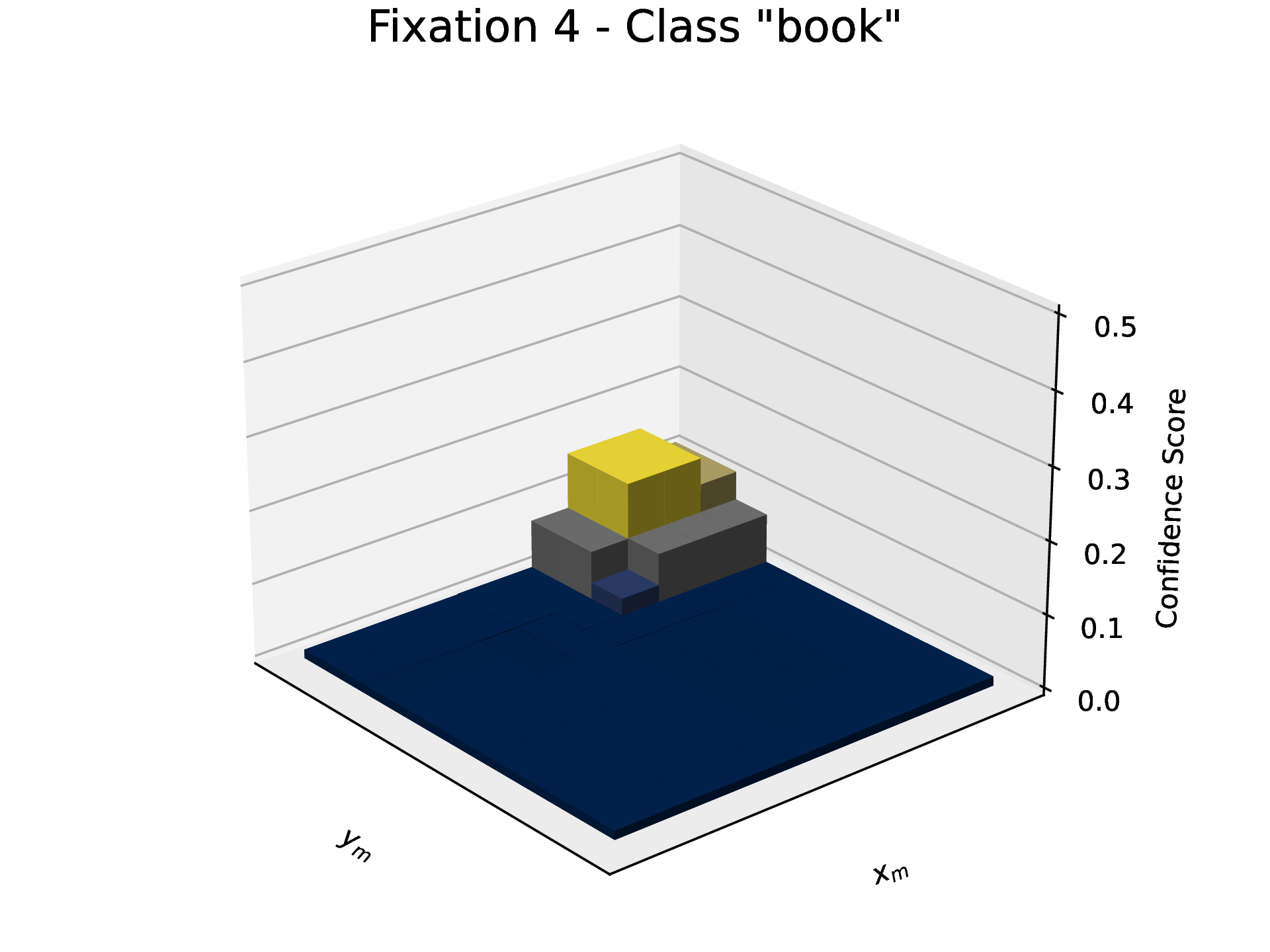}}
    \caption{Example of a visual search experiment, using the full semantic-based methodology, presented in Section \ref{sec:methods}, with foveal uncertainty characterization. The active perception approach applied in this example is fully predictive \eqref{eq:peaks_pred}, selecting the cell that maximizes the class posterior probability \eqref{eq:semanticmap} after simulating a complete map update. The target class, pre-defined for this experiment, is \textit{'book'} and there are multiple instances of this class in the top-right corner of the image that is being used to simulate a visual field. Here we present the YOLOv3 \cite{yolo} object predictions generated at each fixation and the resulting saccade paths (figures in the top), as well as the target class confidence scores (figures at the bottom), for each cell of the map, in the form of bar graphical plots. After only 4 fixations, the semantic-based active perception model is able to direct the gaze toward the region where the target object is actually located.}   
\label{fig:vsexample}
\end{figure*}
The second conclusion is that the active predictive approaches \cite{main} comfortably outperform the non-predictive approaches, regardless of the use of the foveal calibration mechanism presented in Section \ref{sec:fom}. Therefore, predictive approaches \eqref{eq:peaks_pred} can provide added accuracy, at the cost of additional computation.

\subsubsection{Comparison with Saliency-Based and Random Search}
In a second experiment with visual search methods, we compare the proposed active approach (with foveal calibrated scores) with random and saliency-based approaches. Although the most relevant comparison would be with the performance achieved by human subjects, this work aims to make the comparison with a conspicuity model \cite{trad_sal} inspired by the behavior of the early primate visual system. However, previous studies (e.g. \cite{ivsn, objsearch}) have shown that humans tend to generally outperform unguided visual search algorithms.
\begin{figure}
\centering
\includegraphics[width=\columnwidth]{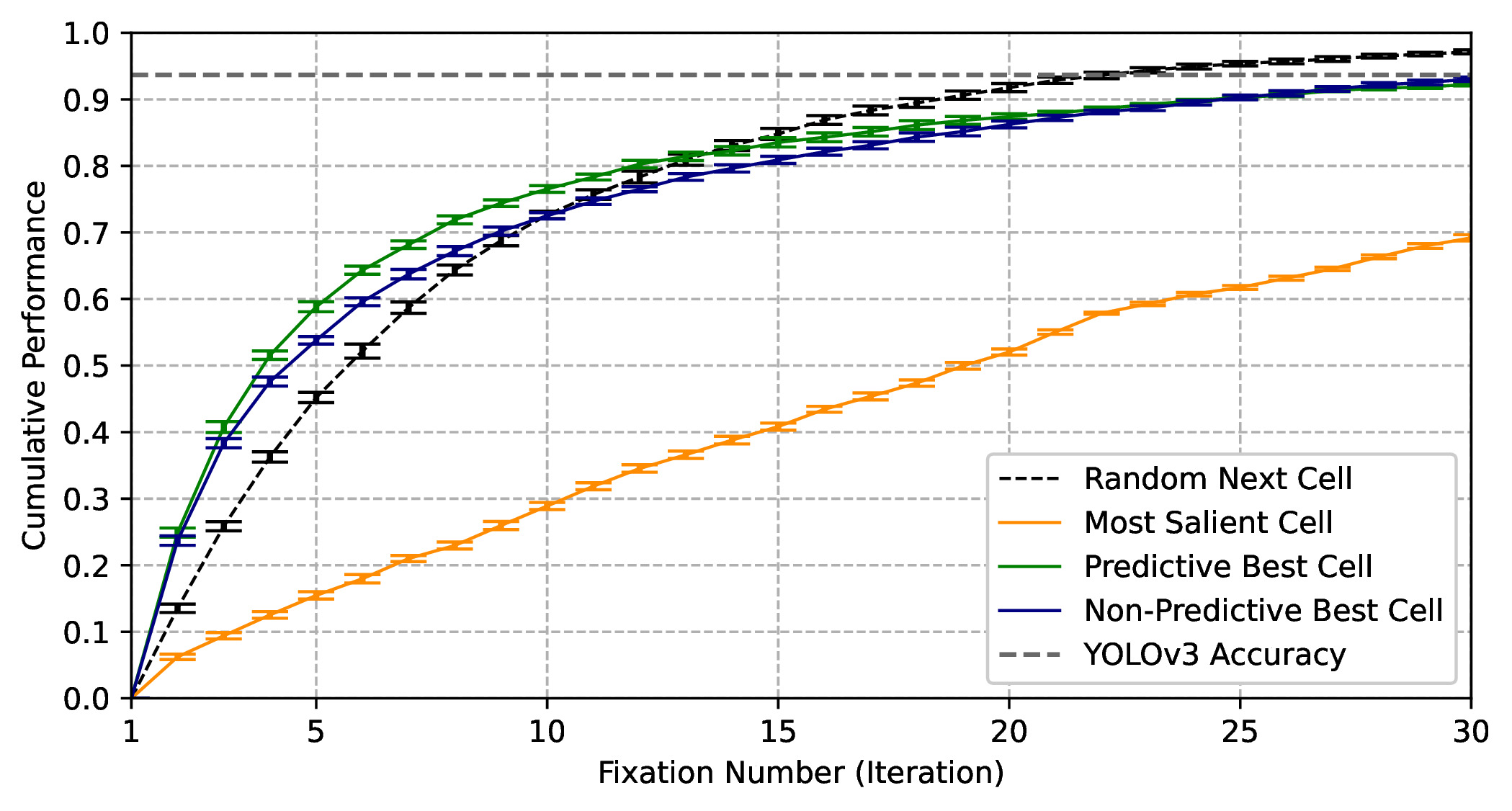}
\caption{Comparison between the mean values of the cumulative performance and the respective SEM bars, obtained using semantic and saliency-based \cite{saliency} approaches, together with the results of the random search. Both predictive and non-predictive models exploit the foveal calibration model.}
\label{fig:cum2}
\end{figure}
The results are displayed in Fig. \ref{fig:cum2} and Tab. \ref{tab2}. From the mean values of the cumulative performance, it is possible to conclude that the semantics-based methods perform better than the unguided random search approach in the initial iterations (up to \nth{10} in the non-predictive mode and up to the  \nth{13} in the predictive mode). We also noticed that the non-predictive approach with non-calibrated scores initially takes the upper hand, yet eventually falls in the face of the random search algorithm.
Regarding the performance of VOCUS2 in object search, it is quite plausible that the lack of precision when finding target objects is directly related to the fact that the search algorithm does not consider class-specific information.
Therefore, the simulated attention mechanism, built on bottom-up saliency information that is extracted from the stimuli at the pixel level, appears to be very inefficient in finding objects in a class-oriented context, in the presence of multiple distractors. 
\begin{table}
\caption{Comparison of Active Visual Search Approaches}
\begin{center}
\begin{tabular}{|c|c|c|c|c|}
\hline
\textbf{Eval. Metric} & \textbf{\textit{Random}}& \textbf{\textit{Saliency}}& \textbf{\textit{Predictive}}& \textbf{\textit{Non-Predictive}} \\
\hline
CP$^{\mathrm{1}}$ @ 5 $\thinspace$  & 45,20\% & 15,47\% & 58,83\% & 53,80\% \\
CP$^{\mathrm{1}}$ @ 15              & 84,80\% & 41,82\% & 84,60\% & 80,90\% \\
CP$^{\mathrm{1}}$ @ 30              & 97,13\% & 69,13\% & 93,10\% & 93,03\% \\
\hline
Max. SEM$^{\mathrm{2}}$   & 1,059\% & 0,627\% & 0,820\% & 0,718\% \\
Comp. Cost$^{\mathrm{3}}$ & 0,015 ms& 762,3 ms& 83,76 ms& 0,133 ms\\
\hline
\multicolumn{5}{l}{$^{\mathrm{1}}$Cumulative Performance. $^{\mathrm{2}}$Standard Error of the Mean. $^{\mathrm{3}}$Per Iteration.}
\end{tabular}
\label{tab2}
\end{center}
\end{table}

\subsubsection{An Illustrative Sample Case}
Let us consider the visual search example, presented in Fig. \ref{fig:vsexample}. The target class is \textit{'book'} and the results illustrate the application of the predictive approach \eqref{eq:peaks_pred} (Section \ref{sec:active}) with the foveal calibration mechanism (Section \ref{sec:fom}).
During the execution of the task, the confidence score of the target class $p_C$ \eqref{eq:confscores} increases consistently, fixation after fixation, in the cells that surround and overlap the location of the target ground truth objects \cite{predvisualfix}, while progressively decreasing in all other cells of \eqref{eq:betamap}. In this example, after only four algorithm iterations (corresponding to 3 complete saccades), the agent would be able to effectively direct its attention toward the region where objects of the category (class) \textit{'book'} are factually located.
Furthermore, notice how YOLOv3 \cite{yolo} is not able to directly detect the targeted instances until the last iteration of the algorithm. 
In this example, the initial focal point was purposely set on a region that is close to the target's opposite corner location, in order to observe the influence of the initial focal point, avoiding direct visibility of the targeted objects at the beginning of the experiment. The observed results align with a fundamental aspect of AET \cite{visualsearch}, which states that, in the context of visual search, the ability to detect and distinguish between target and distractor objects is quite crucial to the human cognitive system. In this example, we observe that the detection of distractors leads the model to actively neglect the regions where these types of objects are contained. This is related to the fact that during the update process \eqref{eq:kaplanrule}  (Section \ref{sec:fusion}) lower target class scores \eqref{eq:scores} tend to decrease the influence of the parameter of \eqref{eq:betamap} associated with that same class, in relation to the parameters associated with higher class scores. Moreover, due to this distinction of target-distractor dissimilarities, the model tends to favor unexplored regions, where the semantic content is yet unknown. The target's initial confidence bar plot, which can be observed in Fig. \ref{fig:vsexample} (yellow cells), shows that the cells that are not overlapped by distractor-related detections present higher target class posteriors \eqref{eq:semanticmap}, in relation to the cells where these types of object are detected. 
Since in this example we apply the predictive approach \eqref{eq:peaks_pred}, the confidences $p_k$ \eqref{eq:confscores} represented in the bar plots do not reflect the estimated confidences that are considered by the active perception model. Hence, to obtain the state \eqref{eq:betamap} of the estimated map that is fed to the optimization function \eqref{eq:peaks_pred}, one must first simulate an update for each cell, using Kaplan's rule \eqref{eq:kaplanrule}, with the estimated local scores $\Bar{S}^{\mathbf{x}}_0$ \eqref{eq:local_scores}, to obtain $\Bar{p}_{t+1,k}$ \eqref{eq:confscores} as described in Section \ref{sec:active}. Therefore, the most promising cell, selected by the active perception mechanism, to where the gaze is shifted, does not correspond to the cell associated with the highest target class probability \eqref{eq:semanticmap}, based on the current state of the semantic map \eqref{eq:betamap}. Finally, observing the evolution of the target's categorical probability \eqref{eq:semanticmap} at different gaze fixations, we conclude that the confidence scores observed in the first iterations are hardly distinguishable, as the absolute differences between them are almost negligible. For the latter fixations, the states of the map clearly differentiate the region where the ground-truth target objects are located from the rest of the image. The target's class confidences start to distance themselves from the confidences associated with the cells that do not cover that same region, therefore validating the proposed methodology.  

\subsection{Scene Exploration}\label{sec:sceneexp}

Regarding exploration experiments, our objective is to capitalize on previously performed experiments \cite{main} to verify how accurately the knowledge collected on the maps represents the semantic content displayed within the confines of the visual scene. Furthermore, we assess the relative computational costs of the different metrics, applied to their respective acquisition functions \cite{apfov}, as presented in Section \ref{sec:results_sceneexp}, to compare the performance of this model with the performances achieved by both saliency-based \cite{saliency} and random gaze selection approaches when it comes to accurately represent the semantic content that is available in the scene.

\begin{figure}
\centering
\includegraphics[width=\columnwidth]{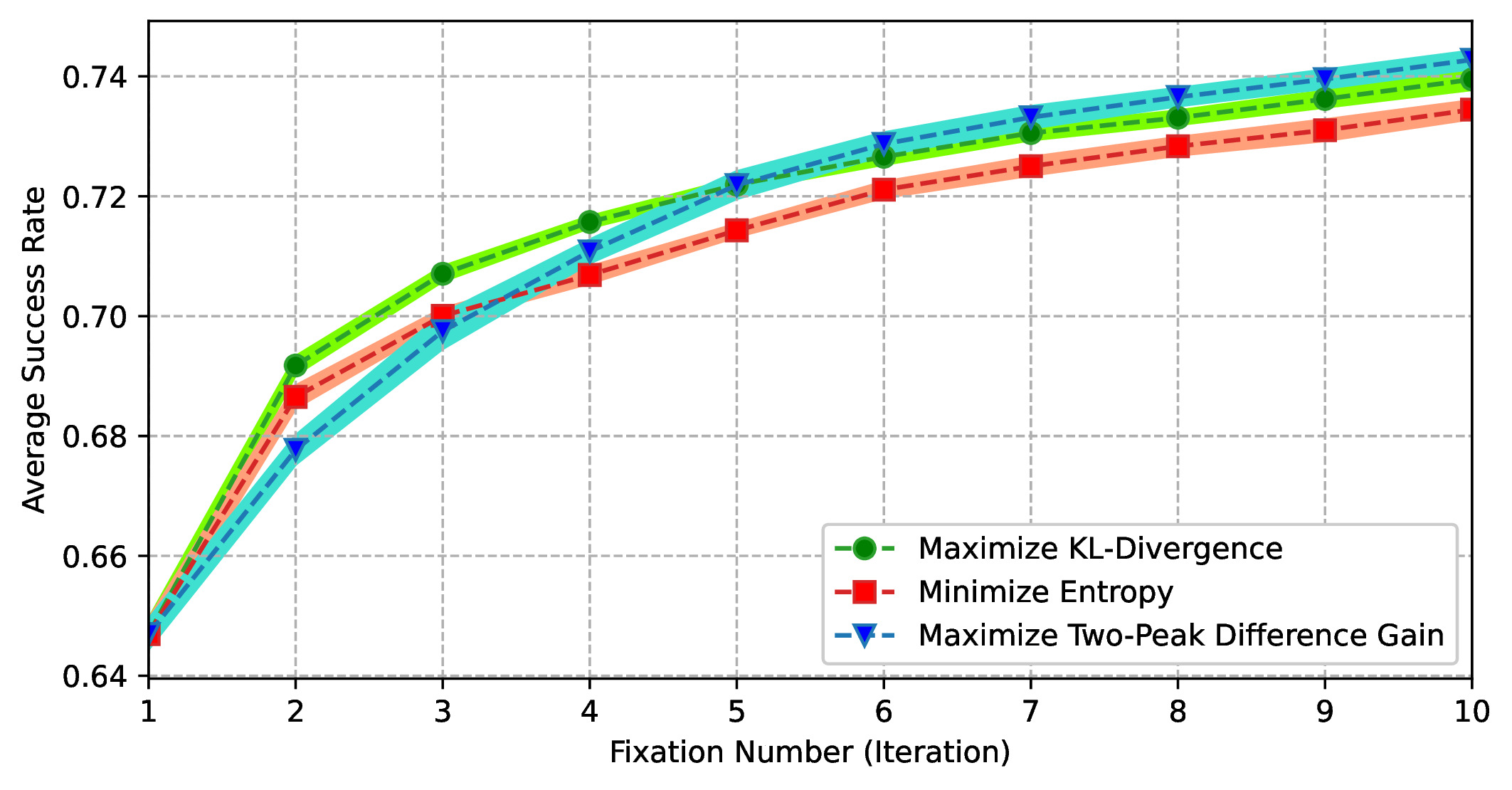}
\caption{Comparison of the mean values of the (average) success rate obtained after 10 repetitions, each starting in a different focal point, with different active perception techniques, together with their respective standard error bands.}
\label{fig:acfuncs}
\end{figure}

First, we performed a comparative analysis of the different acquisition functions, on the map obtained using foveal calibrated scores \eqref{eq:corrected_scores}.
The mean and SEM of the success rates in each saccade, obtained for all images from the validation set, are presented in Fig. \ref{fig:acfuncs}. From the results displayed in Fig. \ref{fig:acfuncs} we elect the maximization of the Kullback-Leibler divergence as the best-performing active perception metric among those considered. This metric \eqref{eq:kldivergence} consistently outperforms negentropy \eqref{eq:entropy} and, in relation to the two-peak difference gain \eqref{eq:twopeaks}, presents a better trade-off between the amount of correct information collected in the first iterations and the long-term performance.

\begin{figure}
\centering
\includegraphics[width=\columnwidth]{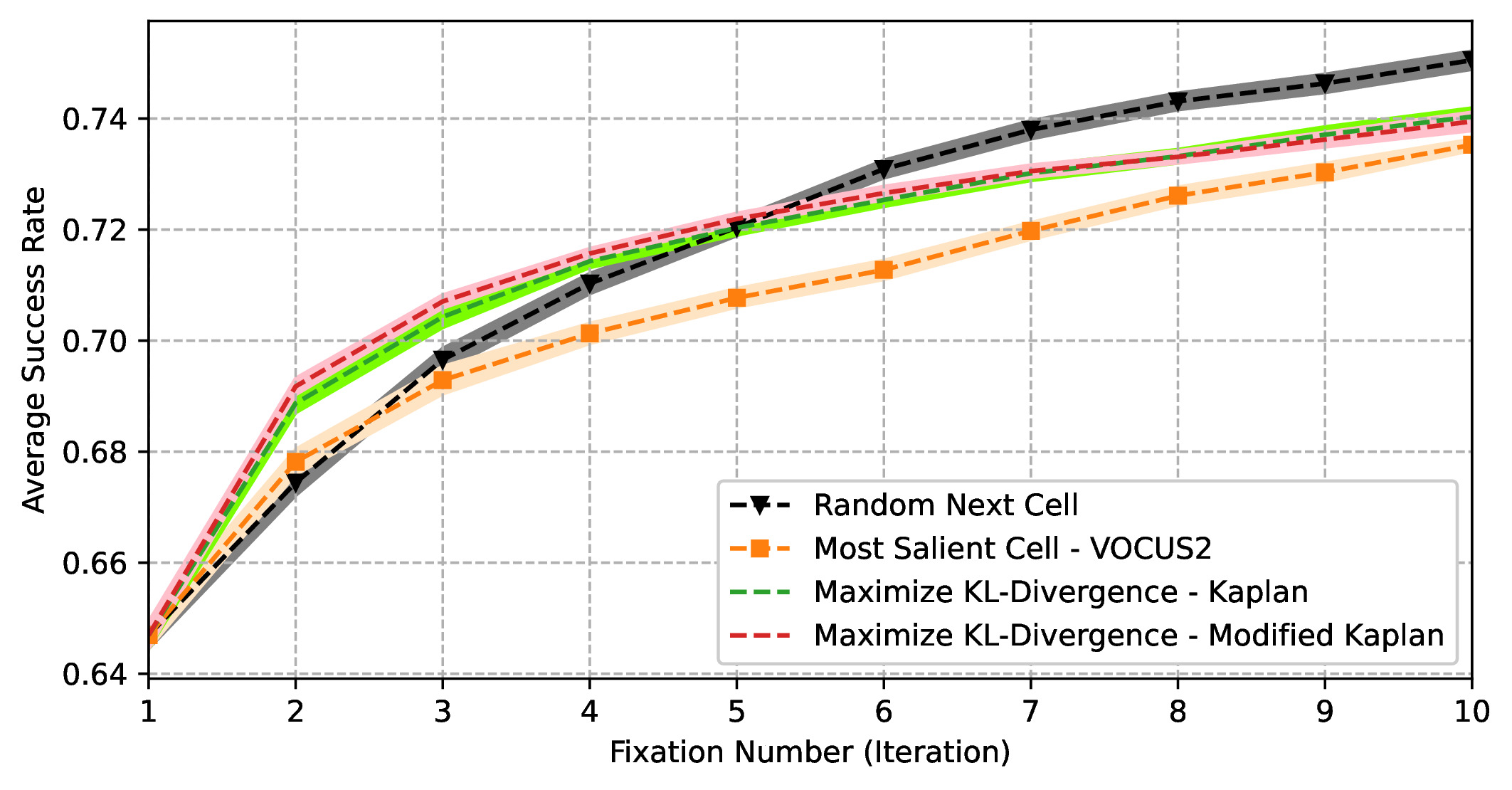}
\caption{Mean values of the average success rate and the respective standard error (SEM) bands, for the semantic-based active perception approach that considers the probability of improving \eqref{eq:minimize} the KL-Divergence metric \eqref{eq:kldivergence} using either the raw (Kaplan) or calibrates (Modified Kaplan) scores, in comparison with both the random and VOCUS2 \cite{saliency} results.}
\label{fig:expres}
\end{figure}

\begin{table}
\caption{Computational Costs$^{\mathrm{1}}$ of Scene Exploration Approaches}
\begin{center}
\begin{tabular}{|c|c|c|c|c|}
\hline
\textbf{\textit{Random}}& \textbf{\textit{Saliency}}& \textbf{\textit{KL-Divergence}}& \textbf{\textit{Entropy}} & \textbf{\textit{Two-Peak}}\\
\hline
\multirow{2}{*}{0,017 ms} & \multirow{2}{*}{0,462 s} 
      & 9.851 s $\thinspace(\mathrm{2})$
      & 9.770 s $\thinspace(\mathrm{2})$ 
      & 8.971 s $\thinspace(\mathrm{2})$\\
      \cline{3-5}  
    & & 9,907 s $\thinspace(\mathrm{3})$
      & 9,833 s $\thinspace(\mathrm{3})$
      & 8,981 s $\thinspace(\mathrm{3})$ \\
\hline
\multicolumn{5}{l}{$^{\mathrm{1}}$Per Iteration. Updates: $^{\mathrm{2}}$w/ Kaplan rule. $^{\mathrm{3}}$w/ Modified Kaplan rule.}
\end{tabular}
\label{tab3}
\end{center}
\end{table}

Secondly, we compare the proposed semantic model \cite{main}, using the Kullback-Leibler divergence metric \eqref{eq:kldivergence}, with a random algorithm, which randomly selects the next gaze direction, and VOCUS2 \cite{saliency}, the elected saliency-based method. Furthermore, we assess the use of calibrated \eqref{eq:corrected_scores} and non-calibrated \eqref{eq:scores} scores in the map update stage.
Fig. \ref{fig:expres} presents the average success rates obtained. By maximizing the Kullback-Leibler divergence, semantic exploration
not only outperforms the saliency-based active exploration but also initially leverages the amount of semantic content that is correctly mapped even in the face of a random selection. Moreover, there is no significant upside in considering the calibrated scores 
rather than the raw class scores. 
After about 4 saccades, the random gaze selection algorithm eventually outperforms the other active perception methods. The reason for this phenomenon is that, after a certain amount of iterations, the fovea has already been placed over all the critical regions of the images, directly capturing the majority of the semantic content that is possible to extract with YOLOv3 \cite{yolo} or any other object detection model. 

Finally, from the data presented in Tab. \ref{tab3}, we draw the conclusion that our implementation of the semantic-based model for exploration\cite{main}, described in Section \ref{sec:methods}, implies a larger computational cost than the saliency-based model. The disparity between the time-cost values obtained with VOCUS2 \cite{saliency} and the proposed methodology \cite{main} comes from the overhead inflicted by the quadratic complexity of the map update process on the exploration algorithm.
The fact that for each individual cell, out of the $X \times Y$ grid, we must
compute a total of $X \times Y$ updates, simulating a gaze shift toward each $(x^\prime,y^\prime)$ cell, is reflected in a computational cost about $20\times$ higher compared to the cost of generating a saliency map. However, the proposed methodology increases the amount of accurately mapped information in the first few saccades.

\section{Conclusion}\label{sec:conclusion}


This work has studied the performance of the semantic model \cite{main} in visual search and exploration tasks, using different active perception metrics, and performing a qualitative and quantitative comparison with VOCUS2 \cite{saliency}, a bottom-up saliency model. We have fully implemented and tested the active perception methodology proposed by Dias et al. \cite{main}, adapting it to the visual search context \cite{visualsearch}. To this end, we designed and meticulously followed the methodological pipeline illustrated in Fig. \ref{fig:general}, fully testing the saliency-based approaches \cite{saliency} and the semantic-based approaches, with and without foveal calibration \cite{uncertainty}, as described in Section \ref{sec:methods}. By using object detection models pre-trained with Cartesian images, it is much faster to fit calibrate the classifier scores with a \cite{dirichlet}-based  foveal observation mechanism within a Bayesian framework \cite{predvisualfix}, than to retrain the detection model with a foveal image dataset.

\subsection{Highlighted Contributions}

Let us start by considering the results obtained during the experimental phase with regard to the scene exploration experiments. First, we have been able to show, through extensive experimentation, that the methodology proposed by Dias et al. \cite{main} is able to beat the performance of a biologically inspired attention mechanism \cite{trad_sal}. Moreover, we have validated the ability of the different active perception techniques presented in Section \ref{sec:results_sceneexp}, when it comes to accurately mapping the available semantic content of the scene. We also wished to understand whether the use of calibrated scores \eqref{eq:corrected_scores}, outperforms the use of direct raw scores \eqref{eq:scores} of YOLOv3 \cite{yolo}. The results of Section \ref{sec:sceneexp} have shown that there is no particular advantage in performing the calibration.

Regarding visual search, the foveal observation model, which calibrates the scores that are used to update a map \eqref{eq:betamap}, has a positive impact on the performance of the semantic model. Furthermore, it is also established that the predictive approach \cite{main}, using the expected value of the probability distribution in each cell to simulate an update, is undoubtedly advantageous, beating the traditional bottom-up saliency model \cite{saliency} by a substantial margin. Notice that VOCUS2 performs significantly under other visual search methods, due to the information considered not being target-oriented. Furthermore, in terms of computational cost, from Tab. \ref{tab2} we infer that, compared to the cost of generating a full conspicuity map, our proposed active perception mechanism for visual search \eqref{eq:peaks_pred} is not only effective, but also quite fast.

Visual search results support the idea that the human cognitive recognition system does not rely solely on bottom-up features \cite{visualsearch}, but also on top-down features and other target similarity principles \cite{wolfe} grounded in the AET. The lack of top-down features in the model \cite{saliency} justifies the inaccuracy exhibited, when completing the search task, in contrast to other recently developed models \cite{ivsn} that incorporate such features.

\subsection{Future Work}

Following the consideration made in Section \ref{sec:active} regarding the greedy nature of the visual search task, it is important not to neglect the semantic information available in the cells surrounding a potential fixation point, which can be extracted from \eqref{eq:predbeta}. It is possible that, by incorporating information available in adjacent cells, both the accuracy and precision of the active perception model increase. However, to achieve spatial precision, weights should be attributed to the information contained in the surrounding cells, according to their distance to the focal point. For future consideration, it would certainly be interesting to investigate whether the integration of information available in adjacent cells could positively influence the performance of the semantic-based active perception model \cite{main}. Such a technique would get more out of the semantic map, taking full advantage of the spatial properties of the cells and the relationships between them, at the cost of slightly increasing the overhead of the algorithm.

Moreover, it would also be interesting to compare the semantic-based model considered in this work \cite{main}, with an adapted version of IVSN \cite{ivsn}, which can iteratively search for stimuli with foveal vision, extracting top-down characteristics along the way. Furthermore, this comparison could also be extended to include novel visual search approaches that are based on deep learning models (e.g. ScanDMM \cite{scandmm}, PathGAN \cite{pathgan}, DeepGaze \rom{3} \cite{kummerer}, HAT \cite{transformers}, or even a model based on a Bayesian ideal observer \cite{gazebayes}) that have delivered better results when it comes to information gain between consecutive saccades and scan path similarity with human subjects. Understand how Large Language models (LLMs) and Visual Language models (VLMs) can provide human knowledge \cite{llm} that can be exploited to make both search and exploration more efficient.

Finally, it would be interesting to adapt the proposed methodology, to perform the same humanoid task, yet with a real mobile agent that must be able to perform body movements \cite{objsearch}, moving through the environment while dynamically changing the range, location, and direction of the field of view. 


\vspace{12pt}


\begin{thebibliography}{00}
\bibitem{foveal} E. E. Stewart, M. Valsecchi, and A. C. Schütz, “A review of interactions between peripheral and foveal vision,” Journal of Vision, vol. 20, no. 12, pp. 2–2, 2020.
\bibitem{activeperception} R. Bajcsy, Y. Aloimonos, and J. K. Tsotsos, “Revisiting active perception,” Autonomous Robots, vol. 42, pp. 177–196, 2018.
\bibitem{activevision} R. P. de Figueiredo and A. Bernardino, “An overview of space-variant and active vision mechanisms for resource-constrained human inspired robotic vision,” Autonomous Robots, pp. 1–17, 2023
\bibitem{odreview} Z. Q. Zhao, P. Zheng, S. -T. Xu and X. Wu, "Object Detection With Deep Learning: A Review," in IEEE Transactions on Neural Networks and Learning Systems, vol. 30, no. 11, pp. 3212-3232, Nov. 2019.
\bibitem{trad_sal} L. Itti, C. Koch and E. Niebur, "A model of saliency-based visual attention for rapid scene analysis," in IEEE Transactions on Pattern Analysis and Machine Intelligence, vol. 20, no. 11, pp. 1254-1259, Nov. 1998.
\bibitem{saliency} S. Frintrop, T. Werner, and G. Martin Garcia, “Traditional saliency reloaded: A good old model in new shape,” in Proceedings of the IEEE Conference on Computer Vision and Pattern Recognition, pp. 82–90, 2015.
\bibitem{topdown} R. Burt, N. N. Thigpen, A. Keil, and J. C. Principe, “Unsupervised foveal vision neural architecture with top-down attention,” Neural Networks, vol. 141, pp. 145–159, 2021.
\bibitem{ivsn} M. Zhang, J. Feng, K. T. Ma, J. H. Lim, Q. Zhao, and G. Kreiman, “Finding any Waldo with zero-shot invariant and efficient visual search,” Nature Communications, vol. 9, no. 1, p. 3730, 2018.
\bibitem{main} A. Dias, L. Simões, P. Moreno and A. Bernardino, "Active Gaze Control for Foveal Scene Exploration," 2022 IEEE International Conference on Development and Learning (ICDL), London, United Kingdom, pp. 115-120, 2022.
\bibitem{predvisualfix} M. Kümmerer and M. Bethge, “Predicting visual fixations,” Annual Review of Vision Science, vol. 9, pp. 269–291, 2023.
\bibitem{uncertainty} M. Xie, S. Li, R. Zhang, and C. H. Liu, “Dirichlet-based Uncertainty Calibration for Active Domain Adaptation,” The Eleventh International Conference on Learning Representations, 2023.
\bibitem{compound} N. L. Johnson, S. Kotz, and N. Balakrishnan, Discrete multivariate distributions, vol. 165. Wiley New York, 1997.
\bibitem{kaplan} L. M. Kaplan, S. Chakraborty and C. Bisdikian, "Fusion of classifiers: A subjective logic perspective," 2012 IEEE Aerospace Conference, Big Sky, MT, USA, pp. 1-13, 2012.
\bibitem{calibration} T. Silva Filho, H. Song, M. Perello-Nieto, R. Santos-Rodriguez, M. Kull, and P. Flach, “Classifier calibration: a survey on how to assess and improve predicted class probabilities,” Machine Learning, vol. 112, no. 9, pp. 3211–3260, Sep. 2023.
\bibitem{visualsearch} L. K. Chan and W. G. Hayward, “Visual search,” Wiley Interdisciplinary Reviews: Cognitive Science, vol. 4, no. 4, pp. 415–429, 2013.
\bibitem{logpolar} V. J. Traver and A. Bernardino, “A review of log-polar imaging for visual perception in robotics,” Robotics and Autonomous Systems, vol. 58, no. 4, pp. 378–398, 2010.
\bibitem{siebert} P. Ozimek, N. Hristozova, L. Balog, and J. P. Siebert, “A space-variant visual pathway model for data efficient deep learning,” Frontiers in Cellular Neuroscience, vol. 13, p. 36, 2019.
\bibitem{grf} D. Pamplona and A. Bernardino, "Smooth Foveal vision with Gaussian receptive fields," 2009 9th IEEE-RAS International Conference on Humanoid Robots, Paris, France, pp. 223-229, 2009.
\bibitem{biofov} H. Lukanov, P. König, and G. Pipa, “Biologically inspired deep learning model for efficient foveal-peripheral vision,” Frontiers in Computational Neuroscience, vol. 15, p. 746204, 2021.
\bibitem{fovsys} A. F. Almeida, R. Figueiredo, A. Bernardino, and J. Santos-Victor, “Deep networks for human visual attention: A hybrid model using foveal vision,” in ROBOT 2017: Third Iberian Robotics Conference: Volume 2, pp. 117–128, 2018.
\bibitem{milicio} C. Melício, R. Figueiredo, A. F. Almeida, A. Bernardino and J. Santos-Victor, "Object detection and localization with Artificial Foveal Visual Attention," 2018 Joint IEEE 8th International Conference on Development and Learning and Epigenetic Robotics (ICDL-EpiRob), Tokyo, Japan, pp. 101-106, 2018.
\bibitem{rcnn} R. Girshick, J. Donahue, T. Darrell, and J. Malik, “Rich feature hierarchies for accurate object detection and semantic segmentation,” in Proceedings of the IEEE conference on computer vision and pattern recognition, pp. 580–587, 2014.
\bibitem{ssd} W. Liu et al., “Ssd: Single shot multibox detector,” in Computer Vision–ECCV 2016: 14th European Conference, Amsterdam, The Netherlands, October 11–14, 2016, Proceedings, Part I 14, pp. 21–37, 2016.
\bibitem{yolo_og} Joseph Redmon, Santosh Divvala, Ross Girshick, Ali Farhadi; Proceedings of the IEEE Conference on Computer Vision and Pattern Recognition (CVPR), pp. 779-788, 2016.
\bibitem{yolo} J. Redmon and A. Farhadi, “Yolov3: An incremental improvement,” University of Washington, 2018.
\bibitem{eva} Y. Fang et al., “Eva: Exploring the limits of masked visual representation learning at scale,” in Proceedings of the IEEE/CVF Conference on Computer Vision and Pattern Recognition, pp. 19358–19369, 2023.
\bibitem{detr} N. Carion, F. Massa, G. Synnaeve, N. Usunier, A. Kirillov, and S. Zagoruyko, “End-to-end object detection with transformers,” in European conference on computer vision, 2020, pp. 213–229.
\bibitem{apfov} R. P. de Figueiredo, A. Bernardino, J. Santos-Victor, and H. Araújo, “On the advantages of foveal mechanisms for active stereo systems in visual search tasks,” Autonomous Robots, vol. 42, pp. 459–476, 2018.
\bibitem{context} A. Torralba, A. Oliva, M. S. Castelhano, and J. M. Henderson, “Contextual guidance of eye movements and attention in real-world scenes: the role of global features in object search.,” Psychological review, vol. 113, no. 4, p. 766, 2006.
\bibitem{saltinet} M. Assens Reina, X. Giro-i Nieto, K. McGuinness, and N. E. O’Connor, “Saltinet: Scan-path prediction on 360 degree images using saliency volumes,” in Proceedings of the IEEE International Conference on Computer Vision Workshops, pp. 2331–2338, 2017.
\bibitem{pathgan} M. Assens, X. Giro-i Nieto, K. McGuinness, and N. E. O’Connor, “PathGAN: Visual scanpath prediction with generative adversarial networks,” in Proceedings of the European Conference on Computer Vision (ECCV) Workshops, pp. 0–0, 2018.
\bibitem{scandmm} X. Sui, Y. Fang, H. Zhu, S. Wang, and Z. Wang, “ScanDMM: A deep markov model of scanpath prediction for 360deg images,” in Proceedings of the IEEE/CVF Conference on Computer Vision and Pattern Recognition, pp. 6989–6999, 2023.
\bibitem{kummerer} M. Kümmerer, M. Bethge, and T. S. Wallis, “DeepGaze III: Modeling free-viewing human scanpaths with deep learning,” Journal of Vision, vol. 22, no. 5, pp. 7–7, 2022.
\bibitem{transformers} Z. Yang, S. Mondal, S. Ahn, G. Zelinsky, M. Hoai, and D. Samaras, “Predicting Human Attention using Computational Attention,” 2023.
\bibitem{objsearch} R. Druon, Y. Yoshiyasu, A. Kanezaki and A. Watt, "Visual Object Search by Learning Spatial Context," in IEEE Robotics and Automation Letters, vol. 5, no. 2, pp. 1279-1286, April 2020.
\bibitem{benchmark} M. Kummerer, T. S. Wallis, and M. Bethge, “Saliency benchmarking made easy: Separating models, maps and metrics,” in Proceedings of the European Conference on Computer Vision (ECCV), pp. 770–787, 2018.
\bibitem{cocosearch18} Y. Chen, Z. Yang, S. Ahn, D. Samaras, M. Hoai, and G. Zelinsky, “Coco-search18 fixation dataset for predicting goal-directed attention control,” Scientific reports, vol. 11, no. 1, p. 8776, 2021.
\bibitem{mit300} T. Judd, F. Durand, and A. Torralba, “A benchmark of computational models of saliency to predict human fixations,” 2012.
\bibitem{cat2000} A. Borji and L. Itti, “Cat2000: A large scale fixation dataset for boosting saliency research,” 2015.
\bibitem{sota} M. Kümmerer and M. Bethge, “State-of-the-art in human scanpath prediction,” 2021.
\bibitem{saccader} G. Elsayed, S. Kornblith, and Q. V. Le, “Saccader: Improving accuracy of hard attention models for vision,” Advances in Neural Information Processing Systems, vol. 32, 2019.
\bibitem{llm} R. Fu, J. Liu, X. Chen, Y. Nie, and W. Xiong, “Scene-LLM: Extending Language Model for 3D Visual Understanding and Reasoning,” arXiv preprint arXiv:2403.11401, 2024.
\bibitem{grotz} M. Grotz, T. Habra, R. Ronsse and T. Asfour, "Autonomous view selection and gaze stabilization for humanoid robots," 2017 IEEE/RSJ International Conference on Intelligent Robots and Systems (IROS), Vancouver, BC, Canada, pp. 1427-1434, 2017.
\bibitem{dirichlet} T. Minka, “Estimating a Dirichlet distribution,” Massachusetts Institute of Technology, Tech Report, 2000.
\bibitem{bishop} C. M. Bishop and N. M. Nasrabadi, Pattern recognition and machine learning, vol. 4. Springer, 2006.
\bibitem{coco} T.Y. Lin et al., “Microsoft COCO: Common Objects in Context,” in Computer Vision – ECCV 2014, pp. 740–755, 2014.
\bibitem{wolfe} J. M. Wolfe, “Visual search: How do we find what we are looking for?,” Annual review of vision science, vol. 6, pp. 539–562, 2020.
\bibitem{gazebayes} S. Rashidi, W. Xu, D. Lin, A. Turpin, L. Kulik, and K. Ehinger, “An active foveated gaze prediction algorithm based on a Bayesian ideal observer,” Pattern Recognition, vol. 143, p. 109694, 2023.
\end{thebibliography}
\end{document}